%% file: acl_latex.tex
\documentclass[11pt]{article}

\usepackage[preprint]{acl}
\usepackage{amsmath}
\usepackage{amssymb}
\usepackage[inline]{enumitem}
\usepackage{times}
\usepackage{latexsym}
\usepackage{booktabs}
\usepackage{multirow}
\usepackage{xcolor}

\usepackage{amsmath}
\usepackage{algorithm}
\usepackage{algpseudocode}
\usepackage{float}
\usepackage{placeins}
\usepackage{listings}

\lstdefinestyle{platour}{
    language=Python,
    basicstyle=\ttfamily\scriptsize,
    keywordstyle=\bfseries,
    commentstyle=\itshape,
    numbers=left,
    numberstyle=\tiny,
    numbersep=5pt,
    frame=single,
    breaklines=true,
    columns=fullflexible,
    keepspaces=true,
    showstringspaces=false,
    captionpos=t
}

\usepackage{tabularx}
\usepackage[T1]{fontenc}
\usepackage[utf8]{inputenc}

\usepackage{microtype}

\usepackage{inconsolata}

\usepackage{graphicx}

\title{On the Opportunities and Challenges of Prefill-to-Autoregressive Cross-Model Transfer}

\author{
  \textbf{Zhenyi Shen\textsuperscript{1*}},
  \textbf{Junru Lu\textsuperscript{2*}},
  \textbf{Wei Liu\textsuperscript{1*}},
  \textbf{Jiangnan Ye\textsuperscript{1}},
  \textbf{Jiazheng Li\textsuperscript{1}},
  \\
  \textbf{Di Yin\textsuperscript{2}},
  \textbf{Xing Sun\textsuperscript{2}},
  \textbf{Tameem Adel\textsuperscript{3}},
  \textbf{Yali Du\textsuperscript{1}},
  \textbf{Lin Gui\textsuperscript{1}},
  \textbf{Yulan He\textsuperscript{1}}
  \\
  \\
  \textsuperscript{1}King's College London, UK
  \\
  \textsuperscript{2}Tencent Youtu Lab, China
  \quad
  \textsuperscript{3}National Physical Laboratory, UK
  \\
  \texttt{\{zhenyi.shen,yulan.he\}@kcl.ac.uk} 
  \\
  \textsuperscript{*}Equal contribution
}

\begin{document}
\maketitle
\begin{abstract}

Prefill–to-Autoregressive Cross-model Transfer (\textbf{PACT}), a form of latent communication between language models, combines a larger model's prompt representations with a smaller model's efficient generation, offering a promising way to balance capability and inference cost. However, existing systems provide limited support for independently evaluating large-small model collaboration, and improved task performance alone does not establish whether the receiver benefits from the sender's information. We introduce \textbf{PLatour}, an open-source PACT framework that connects frozen models through lightweight, zero-initialized linear bridges trained directly with task rewards and supports hybrid-attention architectures. PLatour significantly outperforms open-source baselines on generation tasks and surpasses Mostik's reported results on shared mathematics and coding benchmarks using a substantially smaller sender. However, controlled experiments show that the gains on some tasks can be reproduced by uninformative sender signals, self-communication, or receiver-only training. We find that trainable bridges can act as parameter-efficient adaptation mechanisms without exploiting useful sender information. We therefore propose PACT-Check to quantify how much of a trained PACT system's gain genuinely comes from communication. Finally, we discuss the opportunities and challenges of PACT systems. Code is available at \url{https://github.com/zhenyi4/platour}.

\end{abstract}

\section{Introduction}

\begin{figure*}[t]
  \centering
  \includegraphics[width=\textwidth]{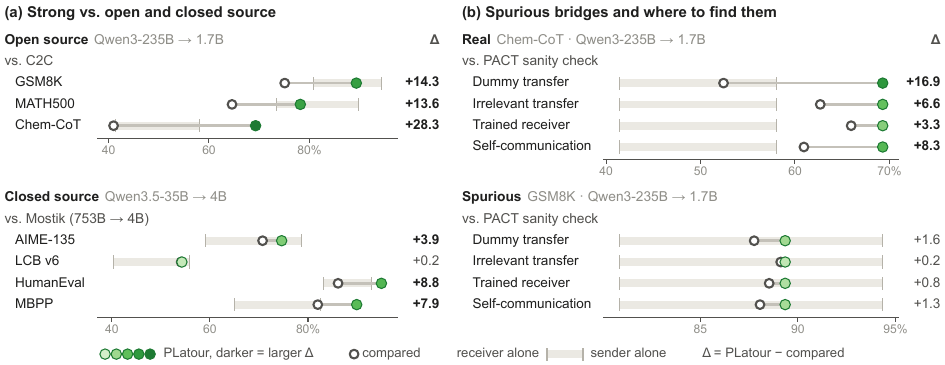}
\captionof{figure}{\textbf{Opportunities and challenges of Prefill–to-Autoregressive Cross-model Transfer (PACT).} \textbf{(a)} PLatour consistently outperforms the open-source C2C \citep{c2c} across all generation tasks and surpasses the closed-source Mostik on math and coding, despite using a much smaller sender. \textbf{(b)} Superior benchmark performance does not necessarily imply effective information transfer from the sender. PACT-Check examines whether these gains depend on sender information through inference-time interventions and separately trained controls. On Chemistry, PLatour substantially outperforms all four sanity checks, providing evidence of meaningful information transfer. In contrast, on GSM8K, all four checks achieve comparable performance, suggesting that PLatour's large gain there cannot be attributed to task-specific information from the sender.}
  \label{fig:overview}
\end{figure*}

Large language models offer strong capabilities but incur substantial computational and memory costs during decoding. Smaller models are more efficient to deploy but often struggle with complex tasks or lengthy instructions. \textbf{PACT} (Prefill–to-Autoregressive Cross-model Transfer), a form of latent communication \citep{mostik, kvcomm,latent_mas,KV-Lingo}, addresses this trade-off by dividing computation between a large \emph{sender} and a small \emph{receiver}. The sender processes the input prompt during prefilling and transfers its internal representations to the receiver, which then generates the response independently. By limiting the sender's involvement to the prefill stage, PACT aims to leverage the capability of larger models while reducing decoding latency. This approach is also compatible with existing prefill--decode disaggregated serving infrastructure \citep{splitwise,distserve}. Recent results demonstrate its promise. For example, \citet{mostik} report closing 47\% of the performance gap between a 753B-parameter sender and a 4B-parameter receiver on mathematical reasoning and coding tasks.

However, Mostik remains closed-source, limiting reproduction and systematic investigation of its approach. Existing open-source methods \citep{c2c,dense_latent,KV-Lingo} provide only a limited understanding of PACT's potential, leaving open questions about how its benefits scale with generation length and sender–receiver capacity gaps, and whether the approach extends to hybrid-attention architectures and larger scales.

Therefore, we introduce \textbf{PLatour} (\textbf{P}refill-only \textbf{LAT}ent c\textbf{O}mm\textbf{U}nication for \textbf{R}easoning), an open-source, simple, and scalable framework for PACT. PLatour directly optimizes the bridged system with reinforcement learning (RL), eliminating the need for the alignment- or distillation-based training commonly adopted in prior work. It connects layers at corresponding relative depths in the sender and receiver through simple, zero-initialized linear projections, avoiding complex bridge architectures. Unlike approaches that couple models through their key–value (KV) caches, PLatour directly transfers hidden states, enabling compatibility with hybrid-attention architectures. Beyond the framework itself, we provide detailed ablation studies of key design choices and share the resulting insights. These designs make PLatour an open and extensible foundation for systematically investigating both the potential and limitations of PACT.

PLatour outperforms existing open-source methods on generation tasks, supporting generation budgets of up to 16K tokens and sender--receiver parameter ratios exceeding 100:1 (Figure~\ref{fig:overview}a). On benchmarks shared with the closed-source Mostik, PLatour surpasses its reported performance on mathematical reasoning and coding tasks, despite using a substantially smaller sender (35B vs.\ 753B parameters). However, further investigation reveals a more nuanced picture. Replacing sender representations with question-independent signals or independently training the receiver with reinforcement learning substantially reduces performance relative to PLatour on some tasks, but achieves comparable performance on others, including some mathematical reasoning settings (Figure~\ref{fig:overview}b). These observations suggest that gains over the standalone receiver may reflect useful sender information, receiver-side adaptation, or both. Performance improvements alone therefore do not establish the contribution of cross-model information transfer. This raises a fundamental question: \textbf{\emph{When do PACT's gains depend on information transferred during prefill?}}

To address this question empirically, we first connect PACT bridges to prefix tuning \citep{prefix} and adapters \citep{peft}, revealing their capacity for adaptation even with frozen backbones. We then introduce PACT-Check, a controlled evaluation protocol that combines interventions on sender information with adaptation baselines. Applied across tasks and training settings, PACT-Check identifies where PACT retains gains beyond the tested controls and where adaptation can reproduce the observed improvements, providing task-dependent evidence of the contribution of sender information. Together, these findings highlight the opportunities and challenges of PACT, while PACT-Check provides a framework for more rigorous evaluation of cross-model latent communication. Our contributions are threefold.

\begin{itemize}
    \item We introduce PLatour, a lightweight hidden-state-bridging PACT framework trained directly with RL, demonstrating its scalability to large sender--receiver capacity gaps, long-form generation scenarios, and hybrid-attention architectures.

    \item PLatour significantly outperforms existing open-source methods on generation tasks and achieves results competitive with the closed-source Mostik while using a much smaller sender, making it a strong baseline. 

    \item Through controlled experiments, we show that comparable gains can arise without useful sender information, challenging their attribution to genuine information transfer. Based on these findings, we propose PACT-Check, an evaluation protocol with four sanity checks, and PACT-Score, which measures the performance gain over the strongest control, to support rigorous evaluation of future PACT systems. We further identify open challenges and opportunities for future research.
\end{itemize}

\section{Related Work}
\label{sec:related_work}
\paragraph{Latent Communication.}

Latent communication enables models to exchange continuous representations, including soft embeddings~\cite{cipher}, intermediate activations~\cite{ac}, and KV caches. Interlat generates and compresses hidden-state messages~\cite{du2026interlat}, while LatentMAS combines autoregressive latent reasoning with KV-memory transfer~\cite{latent_mas}. KVComm reuses caches across agent-specific contexts to reduce redundant prefilling~\cite{kvcomm}. For heterogeneous models, C2C learns task-supervised KV fusion~\cite{c2c}, later made lightweight with a shared low-dimensional bottleneck~\cite{rossi2026lcf}, while other approaches learn shared KV spaces or generation-oriented cache transformations~\cite{kv_alignment,dense_latent}. KV-Lingo and a growing line of cross-model KV transfer methods instead translate caches for model switching without re-prefilling~\cite{KV-Lingo,heo2026closedform,qu2026cachebridge,qiu2026cachelink,li2026universal,yun2026heterofold}, optionally recomputing selected tokens~\cite{sarkar2026rarecache} or porting the recurrent states of hybrid models~\cite{villani2026latentport}; these methods treat the sender's cache as a substitute for the receiver's own prefill and are evaluated by fidelity to it. Within this landscape, PACT restricts the larger sender to prompt encoding and assigns all subsequent autoregressive generation to a smaller receiver, aiming to improve the trade-off between performance and inference efficiency. PLatour follows this division through multi-layer residual injection of sender-prefill hidden states, training the bridge with task rewards. Beyond bridge design, prior work \citep{pact_check1,pact_check2} has investigated whether latent communication systems benefit from sender-provided information through inference-time interventions with model parameters held fixed. PACT-Check focuses on PACT systems with trainable bridges, complementing such interventions with separately trained receiver-only and self-communication controls to assess whether the observed gains can be reproduced through receiver-side adaptation.

\paragraph{Common Representation Space.}
Studies of representational similarity reveal common structure across independently trained networks~\cite{cka}, while the Platonic Representation Hypothesis suggests that increasingly capable models may converge toward shared representational geometry~\cite{plato}. Work on latent-space alignment demonstrates that simple transformations can translate representations between pretrained models~\cite{maiorca2024latentspacetranslationsemantic,pmlr-v243-lahner24a}. Relative representations further enable zero-shot model stitching through similarities to shared anchors~\cite{moschella2023relativerepresentationsenablezeroshot}. For language models, KV cache alignment learns a shared cache representation space to support communication between heterogeneous models~\cite{kv_alignment}. PLatour enables communication between frozen language models by learning only linear projectors that map sender-prefill features to receiver residual updates.

\paragraph{Parameter-Efficient Fine-Tuning.}
Parameter-efficient fine-tuning (PEFT) adapts pretrained models by optimizing a small number of additional or selected parameters while keeping most pretrained weights fixed. Prefix tuning learns continuous prefix activations shared across examples to condition generation~\cite{prefix}, while adapter tuning introduces residual transformations of hidden states~\cite{peft}. PLatour connects to both approaches: trainable question-independent sender signals resemble learned prefixes, while self-communication reduces to a linear residual adapter when sender and receiver use the same hidden state. A unified view connects adapters, prefix tuning, and LoRA as different forms of learned hidden-state modifications~\cite{unified}. This perspective motivates comparing PLatour with receiver-side PEFT baselines such as LoRA~\cite{lora}. Together with sender-independent controls, these comparisons help assess whether bridging gains require useful sender information or can be explained by receiver adaptation.

\input{method_v2}

\section{Experimental Setup}

We evaluate PLatour in two complementary settings. First, we use the Qwen3 family \citep{qwen3} to compare communication interfaces and training objectives under a full-attention architecture, which is compatible with existing KV-based baselines. Second, we use Qwen3.5 \citep{qwen3.5} to evaluate compatibility with hybrid-attention architectures and performance on harder mathematics and coding benchmarks, including comparisons with Mostik's reported results. Across these settings, PACT-Check assesses whether performance gains depend on useful sender information or can be reproduced through receiver-side adaptation.

\paragraph{Datasets.}
We study latent communication on both answer-only and generation tasks. For answer-only tasks, we use ARC-C \citep{arcc}, MMLU \citep{mmlu}, and Chemistry L3 from SciKnowEval \citep{chem}. We use the official ARC-C training and test splits, sample 10,000 MMLU training examples and evaluate on MMLU-Redux \citep{mmlu_relux}, and split Chemistry L3 into training and test sets at a 9:1 ratio. For generation tasks with Qwen3, we use Chemistry \citep{chem}, GSM8K \citep{gsm8k}, and MATH \citep{math}. Chemistry uses the same split as above, with chain-of-thought generation enabled. Mathematical reasoning is evaluated on the official GSM8K test set and MATH500 \citep{verify}. For Qwen3.5, we train separate bridged models for mathematics and coding. The mathematics model is trained on DAPO-17k \citep{dapo} and evaluated on AIME 2022--2026, HMMT 2025 (February and November), and IMO-AnswerBench. Our AIME set comprises 135 problems: AIME I and II from 2022--2025, and AIME I from 2026. We identify 55 AIME problems overlapping with the prepared DAPO data pool: 53 with training and two with validation. We report results on both the full AIME-135 set and the fixed AIME-80 subset excluding all 55 overlaps. The coding model is trained on DeepCoder \citep{deepcoder} and evaluated on LiveCodeBench v6 \citep{lcb}, HumanEval \citep{humaneval}, and MBPP \citep{mbpp}. We deduplicate DeepCoder-24k questions, exclude overlaps with the coding benchmarks and tasks without reliably checkable outputs, and split the remaining 19,074 questions into 18,873 training and 201 validation examples, keeping near-duplicate questions in the same split. All evaluations use non-thinking mode with the model-specific sampling settings recommended by Qwen \citep{qwen3,qwen3.5}.

\paragraph{Baselines.}
We compare PLatour against 12 baselines, grouped into four categories: standalone models, existing communication methods, receiver fine-tuning, and communication controls. Table~\ref{tab:baselines} summarizes their main characteristics.
\begin{enumerate*}[label=\roman*),itemjoin={\quad}]
\item\underline{Standalone models.}
\emph{Receiver Alone} and \emph{Sender Alone} evaluate the off-the-shelf post-trained receiver and sender independently, establishing reference performance without communication.

\item\underline{Existing communication methods.}
We compare against three communication baselines. \emph{C2C} \citep{c2c} projects and fuses sender KV caches into the receiver's caches using SFT. \emph{KV-Lingo} \citep{KV-Lingo} initializes its bridge through closed-form linear regression on 500 training examples, replaces the receiver's KV caches with projected sender caches, and subsequently trains using a KL-divergence objective. \emph{TextComm} provides sender-generated textual guidance, rather than latent representations, to the receiver. The sender decomposes the problem without generating a final solution, and its guidance is concatenated with the original question. Although TextComm requires sender-side generation and therefore falls outside the PACT setting, it provides a reference for comparing textual and latent communication. Further details are given in Appendix~\ref{app:textcomm}.

We additionally include published results from \emph{Mostik} \citep{mostik}, a closed-source latent communication method supporting long-form generation and large sender--receiver size gaps. Its reported results are limited to a frozen GLM-5.2 sender and a frozen Qwen3.5-4B receiver; we therefore treat these as reference results rather than fully controlled comparisons.

\item\underline{Receiver fine-tuning.}
To assess whether performance gains arise from sender communication or receiver adaptation alone, we include \emph{LoRA-SFT}, \emph{Full-SFT}, \emph{LoRA-GRPO}, and \emph{Full-GRPO}. These methods fine-tune the receiver without sender assistance, using either supervised fine-tuning or GRPO. LoRA uses rank $r=128$ unless otherwise specified. Full-parameter fine-tuning uses a learning rate 10 times lower than its LoRA counterpart.

\item\underline{Communication controls.}
\emph{Self-Communication} uses the receiver as both sender and receiver, testing whether gains require cross-model communication. \emph{Virtual Sender} replaces external sender representations with randomly initialized, trainable tensors, testing whether improvements depend on information supplied by an external model.
\end{enumerate*}

\begin{table*}[t]
    \centering
    \footnotesize
    \setlength{\tabcolsep}{6pt}
    \begin{tabular}{@{}l cccc cccc@{}}
        \toprule
        & \multicolumn{4}{c}{Answer-Only / Without CoT}
        & \multicolumn{4}{c}{Generation / With CoT} \\
        \cmidrule(lr){2-5}
        \cmidrule(l){6-9}
        Method & ARC-C & MMLU-Redux & Chemistry & Avg.
        & Chemistry & GSM8K & MATH500 & Avg. \\
        \midrule
        Receiver-Alone
        & 74.23 & 53.96 & 39.05 & 55.75
        & 41.43 & 80.86 & 73.40 & 65.23 \\
        \midrule
        \multicolumn{9}{@{}l}{\textit{Qwen3-235B $\rightarrow$ Qwen3-1.7B}} \\
        C2C
        & 92.75 & \textbf{83.68} & 60.95 & 79.13
        & 41.01 & 75.09 & 64.60 & 60.23 \\
        KV-Lingo
        & 91.21 & 77.88 & 40.95 & 70.01
        & 40.95 & 17.06 & 11.95 & 23.32\\
        PLatour-SFT
        & \textbf{95.56} & 82.90 & \textbf{78.57} & \textbf{85.68}
        & 57.80 & 85.63 & 70.65 & 71.36 \\
        PLatour
        & -- & -- & -- & --
        & \textbf{69.29} & \textbf{89.35} & \textbf{78.2} & \textbf{78.95} \\
        \addlinespace
        \textit{TextComm}
        & 81.83 & 60.25 & 38.10 & 60.06
        & 40.77 & 90.30 & 76.10 & 69.06 \\
        \textit{Sender-Alone}
        & 96.25 & 84.91 & 56.19 & 79.12
        & 58.04 & 94.35 & 89.70 & 80.70 \\
        \midrule
        \multicolumn{9}{@{}l}{\textit{Qwen3-8B $\rightarrow$ Qwen3-1.7B}} \\
        C2C
        & 84.81 & \textbf{72.90} & \textbf{55.71} & 71.14
        & 34.94 & 75.85 & 67.90 & 59.56 \\
        KV-Lingo
        & 86.69 & 70.17 & 43.33 & 66.73
        & 40.48 & 78.35 & 69.85 & 62.89 \\
        PLatour-SFT
        & \textbf{89.85} & 71.29 & 52.86 & \textbf{71.33}
        & 38.69 & 83.62 & 72.10 & 64.80 \\
        PLatour
        & -- & -- & -- & --
        & \textbf{68.39} & \textbf{87.91} & \textbf{76.50} & \textbf{77.60} \\
        \addlinespace
        \textit{TextComm}
        & 80.46 & 58.35 & 37.14 & 58.65
        & 43.27 & 88.13 & 74.40 & 68.60 \\
        \textit{Sender-Alone}
        & 91.72 & 74.61 & 40.48 & 68.94
        & 41.13 & 92.49 & 83.80 & 72.47 \\
        \bottomrule
    \end{tabular}
    \caption{\textbf{Cross-model communication results (accuracy, \%) using Qwen3 senders and a Qwen3-1.7B receiver}. Arrows denote sender $\rightarrow$ receiver. Avg.\ denotes the arithmetic mean of the three scores in each group. Bold scores indicate the best PACT result for each sender. \textit{TextComm and Sender-Alone} are references outside PACT.}
    \label{tab:qwen3_cross_model_transfer}
\end{table*}

\noindent\textbf{Proposed method and variant.}
\emph{PLatour} projects sender hidden states into the receiver's residual stream during prompt processing and optimizes the projectors using GRPO, while keeping both model backbones frozen. \emph{PLatour-SFT} uses the same architecture but trains the projectors with SFT, isolating the effect of the optimization objective.

\noindent\textbf{Training protocol.}
All SFT-based methods except KV-Lingo use labeled responses for answer-only tasks and unfiltered sender-generated rollouts on training questions for generation tasks. By design, KV-Lingo instead trains on the receiver's own rollouts. For models exceeding 100B parameters, PLatour uses their FP8 versions to reduce memory burden. We evaluate PLatour only on generation tasks, as its GRPO training relies on generated responses.

\section{Experimental Results}
We first evaluate whether PLatour consistently outperforms the receiver-alone baseline. We then apply PACT-Check to assess whether these gains depend on useful sender information or can be reproduced through receiver-side adaptation.

\subsection{Does PLatour consistently outperform the receiver-alone baseline?}

%AIME results by year and HMMT results by February and November competition, including additional saved checkpoints, are reported in Appendix~\ref{app:math_breakdowns}.

\paragraph{Answer-Only Tasks.} 
For answer-only tasks, the sender provides information to the receiver at prompt positions preceding the final prompt token, allowing it to directly influence the receiver's prediction. Table~\ref{tab:qwen3_cross_model_transfer} shows that all three cross-model transfer methods substantially improve over the standalone receiver, except KV-Lingo on Chemistry. PLatour-SFT and C2C achieve similar performance, which may reflect limited headroom for further gains on these tasks. KV-Lingo generally performs worse. Further ablations in Appendix~\ref{app:kvlingo_ablation} show that replacing its KL-divergence loss with cross-entropy improves accuracy on all answer-only benchmarks for both sender sizes, suggesting that the KL-based objective contributes to this performance gap.

\paragraph{Generation Tasks.}
Generation tasks require the receiver to generate its own chain of thought without further sender communication, posing a greater challenge for PACT. Across both cross-scale settings, C2C and KV-Lingo underperform Receiver Alone on all three generation benchmarks, whereas PLatour-SFT underperforms it only on MATH500 and on Chemistry with the 8B sender. PLatour-SFT's advantage over C2C arises from both propagation and hidden-state transfer (Appendix \ref{app:hidden_vs_kv}). For hidden-state transfer, it uses a single projector per layer and lets the receiver construct its own keys and values, whereas KV transfer learns separate mappings for keys and values. KV-Lingo slightly outperforms C2C with the 8B sender but degrades substantially on GSM8K and MATH500 with the 235B sender. We attribute KV-Lingo's poor generation performance in the 235B sender setting to its direct KV-cache replacement: the large capacity gap may make the linear-regression initialization of $W$ less effective at aligning the two models' KV representations (Appendix~\ref{app:kvlingo_ablation}). PLatour-SFT also fails to improve over Receiver Alone on MATH500. We hypothesize that imitating longer sender-generated trajectories is difficult for the smaller receiver, a difficulty that also affects receiver-only SFT (Table \ref{tab:confound_qwen3}). In contrast, PLatour consistently outperforms Receiver Alone on all three generation benchmarks under both cross-scale settings. These results favor direct reward maximization over the alignment and distillation objectives in PACT training.

\begin{table*}[t]
    \centering
    \footnotesize
    \setlength{\tabcolsep}{4.5pt}
    \begin{tabular}{@{}ll ccccc cccc@{}}
        \toprule
        &
        & \multicolumn{5}{c}{Mathematics (DAPO-17k)}
        & \multicolumn{4}{c}{Coding (DeepCoder-24k Clean)} \\
        \cmidrule(lr){3-7}
        \cmidrule(l){8-11}
        Method & Sender & AIME-80 & AIME-135 & HMMT & IMO-Ans. & Avg.
        & LCB v6 & HumanEval & MBPP & Avg. \\
        \midrule
        Receiver-Alone & --
        & 59.22 & 59.12 & 46.35 & 41.13 & 48.90
        & 40.36 & 83.16 & 65.08 & 62.87 \\
        \midrule
        PLatour & 397B
        & 71.88 & 72.73 & 61.77 & 49.25 & 60.97
        & 52.50 & 93.67 & 85.43 & 77.20 \\
        PLatour & 35B
        & \textbf{75.63} & \textbf{74.63} & \textbf{64.58} & \textbf{49.75} & \textbf{63.32}
        & \textbf{54.29} & \textbf{94.97} & \textbf{89.93} & \textbf{79.73} \\
        Mostik & 753B
        & -- & 70.74\rlap{$^{*}$} & -- & -- & --
        & 54.14\rlap{$^{*}$} & 86.13\rlap{$^{*}$} & 82.00\rlap{$^{*}$} & 74.09 \\
        \midrule
        \textit{Sender-Alone} & \textit{397B}
        & 85.86 & 86.39 & 82.29 & 61.06 & 76.40
        & 63.79 & 95.43 & 86.78 & 82.00 \\
        \textit{Sender-Alone} & \textit{35B}
        & 80.16 & 78.70 & 69.69 & 55.25 & 68.37
        & 55.79 & 92.99 & 82.48 & 77.09 \\
        \bottomrule
    \end{tabular}
    \caption{\textbf{Mathematical reasoning and code generation accuracy (\%) with a Qwen3.5-4B receiver}. Senders are Qwen3.5 models except for Mostik, which uses GLM-5.2 (753B). AIME-80 excludes the 55 problems overlapping with DAPO-17k; AIME-135 is the full set. IMO-Ans.\ denotes IMO-AnswerBench v2. Math Avg.\ is the unweighted mean of AIME-80, HMMT, and IMO-Ans.; coding Avg.\ is the unweighted mean of the three coding benchmarks. DeepCoder-24k Clean is decontaminated against all three coding test sets. Bold scores indicate the best result in each column, excluding Receiver-Alone and Sender-Alone. $^{*}$Mostik scores are reported results.}
    \label{tab:qwen35_res}
\end{table*}

\paragraph{Scaling.}
To compare with Mostik, we scale PLatour to the Qwen3.5 family, train on more challenging datasets, and evaluate on shared mathematics and coding benchmarks. As shown in Table~\ref{tab:qwen35_res}, PLatour outperforms Mostik's reported results using a substantially smaller sender (35B vs.\ 753B). However, this is not a fully controlled comparison: Mostik's training data are undisclosed, and it is unclear whether it uses a general bridge across task domains, whereas we train separate bridges on domain-specific data. The 397B-sender variant underperforms its 35B counterpart. A single pronounced spike in clipping during training is followed by a substantial and persistent decline in performance, with limited subsequent recovery. Check details in Appendix~\ref{app:clip}.

%(some training instabilty occurs for the 397b model training. will solve them by something like adding norm or things to the bridge so that we can have stable training result. hope to see larger sender always bringing better performance here, though the magnitude of the performance may not justify the cost. also, will talk about why 35b->4b performs worse than grpo here later.)

\begin{figure*}[!t]
    \centering
    % \refstepcounter{figure}
    \includegraphics[width=\textwidth]{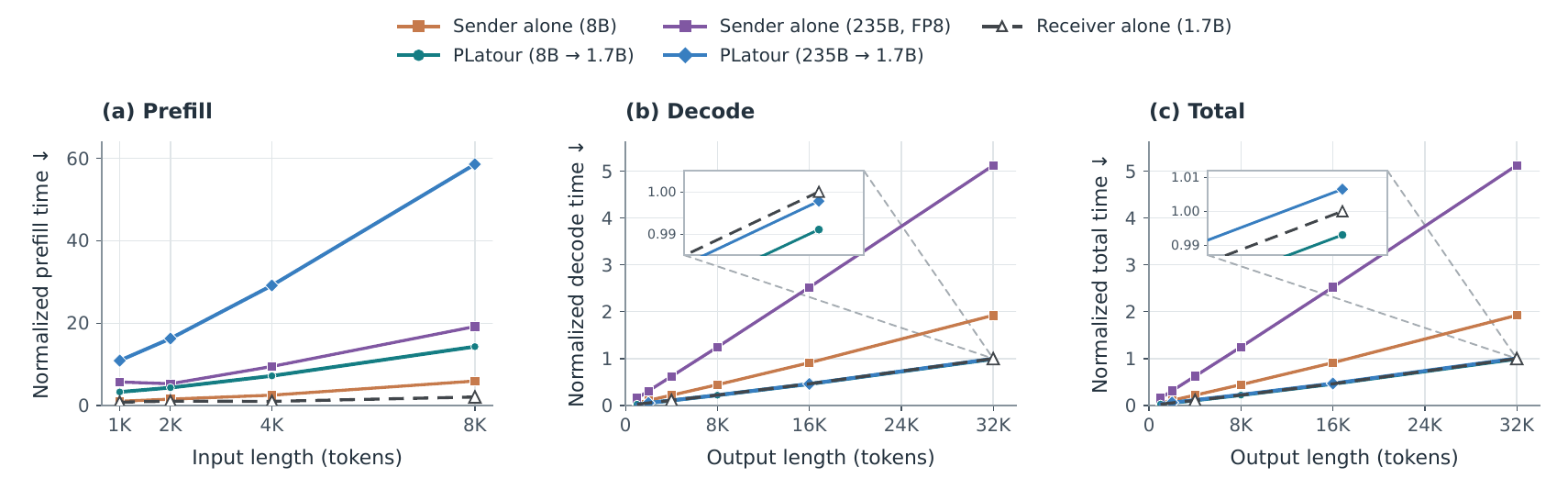}
        \caption{\textbf{Inference efficiency.} (a) Prefill latency, measured as time to first token (TTFT), versus input length. (b) Decoding time and (c) total inference time versus output length, with input length fixed at 4K tokens.}
\label{fig:efficiency}
\end{figure*}

\paragraph{Inference Efficiency.}
Figure~\ref{fig:efficiency} compares prefill latency (TTFT), decoding time, and total inference time. Although PLatour incurs additional upfront latency from prefilling both models and injecting sender representations into the receiver, this fixed cost is amortized over longer generations. As decoding increasingly dominates total latency, PLatour achieves overall inference times close to those of the standalone receiver.

\subsection{Do PLatour's gains depend on useful sender information?}

\begin{figure*}[!t]
    \centering
    \includegraphics[width=\textwidth]{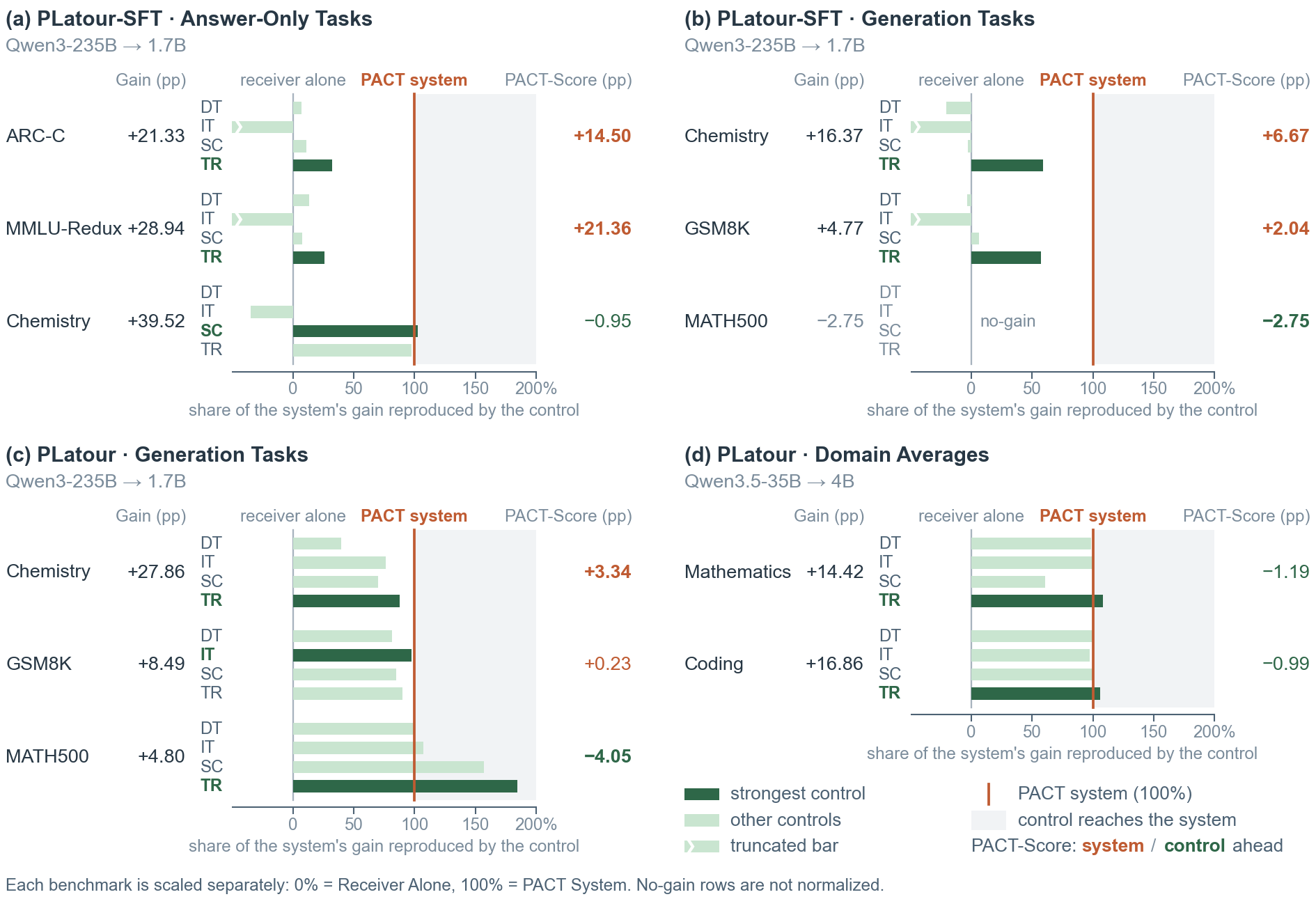}
        \caption{\textbf{PACT-Check results for PLatour and PLatour-SFT}. Panels (a)--(c) show Qwen3 results, and panel (d) shows Qwen3.5 domain averages. Each benchmark is normalized independently, with Receiver Alone at 0\% and the evaluated PACT system at 100\%. Bars show the fraction of the system's gain reproduced by each control (DT, IT, SC, and TR); darker bars mark the strongest control. Gain and PACT-Score report absolute accuracy differences in percentage points: Gain compares the system with Receiver Alone, while PACT-Score compares it with the strongest control. A positive PACT-Score indicates performance beyond all tested controls, providing evidence for a benefit from sender information. In panel (b), TR uses the better of full-parameter SFT and LoRA-SFT for each benchmark to provide an SFT-matched comparison. MATH500 is marked ``no-gain'' because PLatour-SFT underperforms Receiver Alone. Per-benchmark accuracies are reported in Tables~\ref{tab:confound_qwen3} and~\ref{tab:confound_qwen3.5}.}
    \label{fig:pact_check}
\end{figure*}

\paragraph{PACT-Check.}
Figure~\ref{fig:pact_check} reveals substantial variation across controls, PACT systems, and tasks. Overall, improvements over Receiver Alone do not necessarily establish beneficial information transfer from a larger sender. PLatour-SFT is generally more sensitive to DT and IT, suggesting greater dependence on input-specific sender representations. In particular, IT substantially degrades PLatour-SFT performance, whereas PLatour remains more robust. Appendix~\ref{app:norm} shows that GRPO-trained bridges have substantially smaller parameter norms than their SFT-trained counterparts, which may partly explain this difference in sensitivity. SC outperforms Receiver Alone in several settings, demonstrating that bridge training can produce gains without a separate sender. The virtual-sender experiments in Tables~\ref{tab:confound_qwen3} and~\ref{tab:confound_qwen3.5} further investigate this adaptation capacity. TR generally outperforms SC and is the strongest control in most settings, showing that standard receiver fine-tuning provides a strong alternative to cross-model communication. PACT systems' gains over TR are limited on mathematics and coding, and neither Qwen3.5-based PACT system surpasses its compared TR baseline in domain-averaged accuracy.

\paragraph{A Unified View of Communication and Adaptation.}
The performance gains retained under DT and IT in some settings motivate a closer examination of the connection between latent communication and receiver-side adaptation in PACT. DT removes question-dependent sender information, while IT supplies representations from an unrelated question. Both controls reveal that some improvements may not require sender information specific to the receiver's input. One possible explanation is that bridge training learns to exploit question-independent components of sender representations. To illustrate this possibility, consider the limiting case in which the sender representation is replaced by a fixed, question-independent signal $n_{\ell,t}$. Equation~\ref{eq:hidden} becomes
\begin{equation}
\label{eq:fixed_signal}
    \widetilde{h}^R_{\ell,t}
    =
    h^R_{\ell,t}
    +
    a_t W_\ell n_{\ell,t}.
\end{equation}
The trainable term $W_\ell n_{\ell,t}$ then acts as a learned conditioning signal, functionally analogous to prefix tuning \citep{prefix}. Similarly, in the formulation of Eq.~\ref{eq:self_comm}, SC reduces to a linear residual adapter \citep{peft}. LoRA and full-parameter fine-tuning represent alternative forms of receiver-side adaptation. These connections highlight a continuum between cross-model communication and receiver adaptation: the same bridge architecture can potentially exploit sender-specific information or improve receiver computation through learned adaptation, even with both backbones frozen. This dual role underscores the need to distinguish gains attributable to useful information transfer from those achievable through adaptation alone.

\begin{table*}[!t]
    \centering
    \footnotesize
    \setlength{\tabcolsep}{8pt}
    \begin{tabular}{@{}l cccccc@{}}
        \toprule
        & \multicolumn{3}{c}{Answer-Only / Without CoT}
        & \multicolumn{3}{c}{Generation / With CoT} \\
        \cmidrule(lr){2-4}
        \cmidrule(l){5-7}
        Method & ARC-C & MMLU-Redux & Chemistry
        & Chemistry & GSM8K & MATH500 \\
        \midrule
        Sender-Alone gain (235B)
        & $+22.02$ & $+30.95$ & $+17.14$
        & $+16.61$ & $+13.49$ & $+16.30$ \\
        \midrule
        \multicolumn{7}{@{}l}{
            \textit{Gain over Receiver-Alone (pp)}
        } \\
        C2C
        & $+18.52$ & $+29.72$ & $+21.90$
        & $-0.42$ & $-5.01$ & $-5.50$ \\
        KV-Lingo
        & $+16.98$ & $+23.92$ & $+4.28$
        & $-0.48$ & $-2.51$ & $-3.55$ \\
        PLatour-SFT
        & $+21.33$ & $+28.94$ & $+39.52$
        & $+16.37$ & $+4.77$ & $-1.30$ \\
        PLatour
        & -- & -- & --
        & $+27.86$ & $+8.49$ & $+4.80$ \\
        \midrule
        \multicolumn{7}{@{}l}{
            \textit{PACT-Score (pp)}
        } \\
        C2C
        & $+11.69$ & $+22.14$ & $-16.67$
        & $-24.94$ & $-12.68$ & $-14.35$ \\
        KV-Lingo
        & $+10.15$ & $+16.34$ & $-34.29$
        & $-25.00$ & $-10.18$ & $-12.40$ \\
        PLatour-SFT
        & $+14.50$ & $+21.36$ & $-0.95$
        & $-8.15$ & $-2.90$ & $-10.15$ \\
        PLatour
        & -- & -- & --
        & $+3.34$ & $+0.23$ & $-4.05$ \\
        \bottomrule
    \end{tabular}
    \caption{\textbf{Gains over Receiver-Alone and PACT-Scores, both in percentage points (pp), with a Qwen3-1.7B receiver.} The first row reports the gain of the strongest standalone sender (Qwen3-235B) over Receiver Alone. For each method and benchmark, we select the sender (235B or 8B) yielding the highest original cross-model accuracy and compute both metrics for that setting. Gain is the selected accuracy minus Receiver-Alone accuracy. PACT-Score subtracts the strongest applicable control among Dummy Transfer (DT), Irrelevant Transfer (IT), Trained Receiver (TR), and Self-Communication (SC), using DT and IT from the selected sender setting and method-matched SC where available. TR uses the better of full-parameter SFT and LoRA-SFT for answer-only tasks, and the better of full-parameter GRPO and LoRA-GRPO for generation tasks, consistently across methods. Positive PACT-Scores indicate performance above all included controls. Unreported results are denoted by --.}
    \label{tab:pact_score}
\end{table*}

\paragraph{PACT-Score.}
Table~\ref{tab:pact_score} contrasts gains over Receiver Alone with PACT-Scores. Accounting for the strongest controls substantially reduces the apparent advantages, particularly on generation tasks. On ARC-C and MMLU-Redux, all three evaluated methods retain substantial positive PACT-Scores. On answer-only Chemistry, however, C2C and KV-Lingo receive strongly negative scores, while PLatour-SFT is close to zero. Although PLatour-SFT's gain exceeds the standalone sender's gain, it does not exceed the strongest control. Thus, even outperforming the sender does not by itself establish beneficial cross-model information transfer. On generation tasks, neither C2C nor KV-Lingo improves over Receiver Alone on any benchmark. Both methods have substantially negative PACT-Scores, indicating performance below the strongest controls. PLatour-SFT improves over Receiver Alone on Chemistry and GSM8K, but its negative PACT-Scores show that receiver-only training achieves larger gains. PLatour retains a positive PACT-Score on Chemistry, providing evidence of benefits beyond the tested controls, whereas its GSM8K score is numerically close to zero. On MATH500, PLatour improves over Receiver Alone but underperforms the strongest control. These results highlight why benchmark gains alone are insufficient to interpretably establish the contribution of sender information.

\FloatBarrier
\section{Discussion}
\label{sec:discussion}
PLatour's low PACT-Scores on several long-form generation tasks suggest that much of its observed improvement can be reproduced through adaptation without useful sender information. These findings highlight a central challenge for PACT: achieving gains that depend on cross-model communication. Nevertheless, our results also identify settings where communication provides benefits beyond the tested controls, motivating further exploration.

\paragraph{Challenges.} Realizing effective prefill-time communication in PACT requires distinguishing knowledge transfer from receiver adaptation. Because a trainable bridge introduces additional parameters into the receiver, reward-driven training does not inherently distinguish between improvements arising from transmitted information and those arising from the bridge's own learned representations. When the receiver can improve on its own, the bridge may instead function as an adapter or a learned prefix, reducing the sender to an expensive source of noise. A central challenge is therefore to develop training objectives that incentivize the receiver to exploit information from the sender rather than achieving gains through adaptation alone. Equally important is designing evaluations that can reliably identify genuine transfer. This requires tasks on which the receiver has sufficient capability to exploit the transmitted information but still has substantial room for improvement and on which the sender possesses a meaningful knowledge advantage, as well as careful control of test-data contamination. Otherwise, apparent gains may reflect adaptation rather than transfer, while genuine transfer may be obscured by the receiver's limitations or benchmark saturation. Further challenges include generalizing PACT across model families with heterogeneous representation spaces, maintaining stable training as sender models scale, and developing mechanistic explanations of what information the bridge transmits and how the receiver uses it.

\paragraph{Opportunities.}
Positive PACT-Scores in some settings suggest that even simple linear bridges can facilitate useful cross-model information transfer. These findings are consistent with the possibility of shared representational structure, as suggested by the Platonic Representation Hypothesis \citep{plato}, but do not establish the hypothesis itself. Extending our evaluation beyond models from the same family would provide a stronger test of this possibility. Beyond strong-to-weak transfer, latent communication may also benefit models of similar capacity or enable weak-to-strong transfer when their capabilities are complementary. Let $A$ and $B$ denote the sets of problems solved by the receiver and sender, respectively. Problems in $B \setminus A$ represent potential opportunities for transfer, even when the sender is weaker overall. The union $A \cup B$ describes potential combined coverage, rather than a guaranteed outcome of communication. Realizing this potential requires the sender's complementary capabilities to be encoded in representations that the receiver can effectively exploit. 
Another opportunity is to explore the independent scaling of problem understanding and solution search. In PACT, we view large-model prefilling as problem understanding and small-model generation as solution search. PLatour offers an early realization of this separation. This suggests two complementary ways to scale reasoning: using a larger model for better problem understanding, and allowing a smaller model to search longer at lower cost. Studying these two directions could reveal new ways to push the capability–cost Pareto frontier.
A further opportunity is to amortize sender computation across requests sharing a long system prompt. For example, an agent harness may contain tens of thousands of tokens of reusable instructions. Sender representations of this shared prefix could be precomputed, cached, and injected during receiver prefilling, avoiding repeated sender computation for each request. Whether such prefix-only communication retains the benefits of question-specific transfer remains an empirical question. Finally, a shared latent interface could offer a complementary communication channel across model families. One possible approach is to select intermediate layers from different models and learn projections into a common representation space. Studying which representational structures can be aligned while preserving task-relevant information could improve our understanding of model representations and inform the development of more general and efficient latent communication methods.

\section{Conclusion}
We formulated Prefill–to-Autoregressive Cross-model Transfer (PACT) as a distinct form of latent communication in which a large sender only prefills the prompt and a small receiver generates the response independently, aiming to achieve Pareto-optimal trade-offs between efficiency and performance through large–small model collaboration. We also introduced PLatour, an open-source PACT framework that serves as a strong baseline. PLatour outperforms existing open-source methods on generation tasks and surpasses the reported results of the closed-source Mostik on shared math and coding benchmarks with a much smaller sender. However, outperforming existing methods does not necessarily mean that the receiver benefits from information provided by the sender. To distinguish genuine information transfer from receiver adaptation, we propose PACT-Check, a suite of four sanity checks that test whether PACT performance depends on question-specific information from the sender. These checks reveal substantial differences across tasks and training objectives: some gains depend on useful sender information, whereas others can largely be reproduced without it. Our findings highlight task adaptation through the trainable bridge as a confounding factor in PACT systems, underscoring the need to distinguish effective information transfer from gains due to adaptation. We hope PLatour and PACT-Check will facilitate more rigorous evaluation and motivate future research on more robust and effective PACT systems.

\section*{Acknowledgments}

This work was supported in part by the UK Engineering and Physical Sciences Research Council through a Turing AI Fellowship (grant no. EP/V020579/1, EP/V020579/2) and the Prosperity Partnership scheme (grant no. UKRI566). ZS is supported by a PhD studentship provided by the Chinese Scholarship Council.


\bibliography{custom}
\clearpage
\onecolumn
\raggedbottom

\appendix

\setcounter{table}{0}
\renewcommand{\thetable}{A\arabic{table}}

\setcounter{figure}{0}
\renewcommand{\thefigure}{A\arabic{figure}}

\section{Implementation Details}

\subsection{PLatour}
Listing~\ref{lst:platour} presents minimal PyTorch-style pseudocode for PLatour's forward pass and generation. Both backbones are frozen and remain in evaluation mode. Each receiver layer uses a zero-initialized, bias-free linear projection, with sender--receiver layer mappings specified for each experimental setting. Projected sender states are added before the receiver block's RMSNorm, only at prompt positions preceding the final token.

Projection parameters and computations use FP32 to preserve small updates and reduce rounding error; projected residuals are cast to the receiver's BF16 dtype before addition. During training, gradients propagate through the frozen receiver to the projections. The actual implementation combines Hugging Face prefill with vLLM decoding through hybrid-cache transfer. Further details are available in our open-source code.

\begin{center}
\begin{minipage}{0.90\linewidth}
\begin{lstlisting}[
    style=platour,
    caption={PLatour forward pass and generation.},
    label={lst:platour}
]
import torch

def forward(platour, prompt, tokens=None, use_cache=False):
    tokens = prompt if tokens is None else tokens
    with torch.no_grad():
        hidden = platour.teacher.model(
            prompt, output_hidden_states=True,
            use_cache=False).hidden_states
        features = platour.teacher_features(hidden)

    # Add projected sender states at receiver block inputs.
    mask = torch.ones_like(prompt, dtype=torch.bool)
    mask[:, -1] = False
    with platour.inject(features, mask):
        return platour.student(tokens, use_cache=use_cache)

@torch.no_grad()
def generate(platour, prompt, eos_id, max_tokens=1024):
    output = forward(platour, prompt, use_cache=True)
    response = []
    for _ in range(max_tokens):
        probs = output.logits[:, -1].float().softmax(-1)
        token = torch.multinomial(probs, num_samples=1)
        response.append(token)
        if token.item() == eos_id or len(response) == max_tokens:
            break

        # Continue with the receiver alone.
        output = platour.student(
            token, past_key_values=output.past_key_values,
            use_cache=True)
    return torch.cat(response, dim=1)
\end{lstlisting}
\end{minipage}
\end{center}

\subsection{C2C}
\label{app:c2c}

We build on the official C2C implementation%
\footnote{\url{https://github.com/thu-nics/C2C}}. Both backbones are frozen and remain in evaluation mode. The final $n$ sender layers are aligned with the $n$ receiver layers. At each aligned layer, separate key and value networks take concatenated sender and receiver cache features and produce gated residuals added to the receiver's native cache. Communication covers the prompt excluding its final token.

The communication modules use BF16, random initialization, hidden and intermediate dimensions of 1,024, and dropout of 0.1. Only these modules are optimized, using cross-entropy over response tokens. Training uses Gumbel-sigmoid gates with temperature annealed from 1 to 0.001; inference uses deterministic binary gates. For generation, the fused prefix cache is transferred to vLLM, which processes the final prompt token and decodes the response.

Within our Transformers/vLLM integration, we compare an unpadded, sample-wise reference adapter with batched adapters using intermediate padding and tail-only padding. The official C2C implementation already supports batching; this comparison concerns our runtime adapters, all of which retain the same C2C fusion architecture. The tail-only adapter keeps prompt and response tokens contiguous and replays native receiver prefix states at each layer to preserve the reference cache semantics. We check numerical consistency against the sample-wise adapter, including FP32 loss and gradient comparisons.

Table~\ref{tab:c2c_padding} reports the GSM8K comparison under matched data and training settings. In this single-run comparison, tail-only padding achieves accuracy close to the sample-wise reference and higher than the intermediate-padding variant.

\begin{table}[H]
\centering
\small
\setlength{\tabcolsep}{10pt}
\begin{tabular}{lrr}
\toprule
Runtime adapter & GSM8K avg@2 & Truncation rate \\
\midrule
Sample-wise, unpadded          & 77.29 & 0.38 \\
Batched, intermediate padding  & 76.42 & 0.30 \\
Batched, tail-only padding     & 77.33 & 0.23 \\
\bottomrule
\end{tabular}
\caption{Comparison of our C2C runtime adapters with Qwen3-4B$\rightarrow$1.7B on GSM8K. All values are percentages. Runs share the same data, split, layer mapping, three-epoch schedule, batch size of 64, learning rate of $10^{-4}$, and response limit of 2,048 tokens.}
\label{tab:c2c_padding}
\end{table}

\subsection{KV-Lingo}
\label{app:KV-Lingo}

We implement the linear, head-mixing variant of KV-Lingo~\cite{KV-Lingo}. Both backbones are frozen and remain in evaluation mode. The final $n$ sender layers are aligned with the $n$ receiver layers. Each receiver layer has separate bias-free linear maps for keys and values, allowing information to mix across heads. Keys are translated before key normalization and rotary positional embedding (RoPE); the receiver's own key normalization and RoPE are applied afterward. Translation covers the prompt excluding its final token.

Training consists of two stages. First, we initialize the maps by closed-form least-squares fitting on 500 training prompts, excluding held-out examples. We accumulate sufficient statistics in FP64 and solve using a pseudoinverse with relative tolerance $10^{-8}$. Translator parameters and projection computations use FP32, with translated states cast to the receiver's BF16 dtype.

Second, we optimize only the translator for three epochs using full-vocabulary KL divergence, $D_{\mathrm{KL}}(p_{\mathrm{native}}\|p_{\mathrm{translated}})$, between the receiver's continuation distributions under its native and translated caches. Both paths use the same teacher-forced continuation. The loss is averaged over continuation tokens within each example and then over examples; no cross-entropy term is added. We use AdamW with learning rate $10^{-4}$, batch size 64, zero weight decay, 5\% warmup followed by cosine decay, and gradient-norm clipping at 1.

For generation, the receiver processes the final prompt token using the translated prefix cache and continues autoregressively. Our shared evaluation pipeline transfers translated caches to vLLM for decoding. The implementation and task-specific configurations are available in our open-source code.

\section{Experimental Details}
\label{app:experimental_details}

\subsection{Datasets and Splits}
\label{sec:datasets}

We train Qwen3 on ARC-C, MMLU, Chemistry (under both CoT and no-CoT settings), GSM8K, and MATH, and Qwen3.5 on DAPO and DeepCoder. Table~\ref{tab:dataset_splits} summarizes the data partitions. For Qwen3 supervised experiments, we reserve approximately 1\% of each training pool as an internal holdout using seed 42 and reuse the question partitions across all methods.

\paragraph{ARC-C.}
We use the official ARC-Challenge splits~\citep{arcc}, with 1,119 training and 1,172 test questions. The training set is divided into 1,107 training and 12 held-out examples. The official 299-example validation set remains separate from this internal holdout. Supervision uses the correct answer option without a rationale.

\paragraph{MMLU.}
We select 10,000 unique questions from the MMLU auxiliary training set~\citep{mmlu}, excluding normalized question overlaps with MMLU-Redux 2.0 and prompts exceeding 1,024 tokens. Evaluation uses MMLU-Redux 2.0~\citep{mmlu_relux}: we apply available label corrections and exclude entries marked \texttt{no\_correct\_answer} or \texttt{expert}, retaining 5,632 questions across 57 subjects. Supervision uses the correct answer option without a rationale.

\paragraph{Chemistry.}
We use the SciKnowEval Chemistry L3 subset~\citep{chem} with the 1,890/210 train/test partition distributed with SDPO~\citep{sdpo}. The CoT and no-CoT settings share the same question partitions. No-CoT supervision uses the correct answer option, whereas CoT supervision uses one sender-generated reasoning trajectory per training question.

\paragraph{GSM8K and MATH.}
For GSM8K~\citep{gsm8k}, we use the official 7,473 training and 1,319 test questions. For MATH~\citep{math}, we use 7,500 training problems and evaluate on MATH500~\citep{verify}. Supervised training uses one sender-generated solution per question. Targets are prepared separately for each sender while question partitions remain fixed; incorrect and length-truncated trajectories are retained.

\paragraph{DAPO.}
We use DAPO-Math-17k~\citep{dapo}. After deduplicating normalized prompts and excluding seven questions with conflicting labels, we retain 17,391 unique problems. We reserve 500 problems for validation using seed 42 and train on the remaining 16,891. External evaluation uses AIME (135 problems), HMMT 2025 (60), and IMO-AnswerBench v2 (400). We identify 55 AIME problems overlapping with DAPO and additionally report results on the remaining 80-problem AIME subset after removing these overlaps.

\paragraph{DeepCoder.}
We use DeepCoder-Preview~\citep{deepcoder}, excluding problems with unsupported checker contracts. The prepared partition contains 18,873 training and 201 validation problems. External evaluation uses the held-out 175-problem LiveCodeBench v6 increment~\citep{lcb}, HumanEval (164 problems)~\citep{humaneval}, and MBPP (500)~\citep{mbpp}.

\begin{table}[htbp]
    \centering
    \small
    \begin{tabular}{lrrr}
        \toprule
        Dataset & Train & Holdout & Test \\
        \midrule
        \multicolumn{4}{l}{\textit{Qwen3}} \\
        ARC-C              & 1,107 & 12  & 1,172 \\
        MMLU               & 9,900 & 100 & 5,632 \\
        Chemistry (no-CoT) & 1,871 & 19  & 210   \\
        Chemistry (CoT)    & 1,871 & 19  & 210   \\
        GSM8K              & 7,398 & 75  & 1,319 \\
        MATH               & 7,425 & 75  & 500   \\
        \midrule
        \multicolumn{4}{l}{\textit{Qwen3.5}} \\
        DAPO               & 16,891 & 500 & --- \\
        DeepCoder          & 18,873 & 201 & --- \\
        \bottomrule
    \end{tabular}
    \caption{Data partitions. Qwen3 rows show supervised training splits; Qwen3 GRPO uses the complete source training pools. Qwen3.5 rows show GRPO training splits. DAPO and DeepCoder use the external test benchmarks described in the text.}
    \label{tab:dataset_splits}
\end{table}

\subsection{Baselines}
\label{app:baselines}

Table~\ref{tab:baselines} summarizes all baseline methods along three dimensions: the information provided by the sender, the trainable components, and the training objective. These distinctions help clarify whether performance improvements arise from transferring information between models, adapting the receiver, or the choice of optimization method. In particular, receiver fine-tuning baselines assess whether similar gains can be achieved without sender assistance, while communication controls test whether meaningful information from an external sender is necessary. PLatour-SFT further isolates the contribution of GRPO by using the same communication architecture with a different training objective.

\begin{table}[htbp]
\centering
\small
\setlength{\tabcolsep}{5pt}
\renewcommand{\arraystretch}{1.12}
\begin{tabularx}{\textwidth}{
    l l l >{\raggedright\arraybackslash}X
}
\toprule
\textbf{Method} &
\textbf{Sender information} &
\textbf{Trainable components} &
\textbf{Training objective} \\
\midrule

\multicolumn{4}{l}{
\textit{Standalone models}
} \\
Receiver Alone &
None &
None &
--- \\
Sender Alone &
--- &
None &
--- \\

\midrule
\multicolumn{4}{l}{
\textit{Existing communication methods}
} \\
C2C \citep{c2c} &
KV caches &
Communication modules &
SFT \\

KV-Lingo \citep{KV-Lingo} &
KV caches &
Bridge &
Linear alignment + KL-based training \\

TextComm &
Textual guidance &
None &
--- \\

Mostik \citep{mostik} &
Latent representations &
Bridge &
As reported in prior work \\

\midrule
\multicolumn{4}{l}{
\textit{Receiver fine-tuning}
} \\
LoRA-SFT &
None &
Receiver LoRA &
SFT \\

Full-SFT &
None &
Full receiver &
SFT \\

LoRA-GRPO &
None &
Receiver LoRA &
GRPO \\

Full-GRPO &
None &
Full receiver &
GRPO \\

\midrule
\multicolumn{4}{l}{
\textit{Communication controls}
} \\
Self-Communication &
Receiver hidden states &
Projectors &
GRPO \\

Virtual Sender &
Learnable virtual states &
Projectors and virtual states &
GRPO \\

\midrule
\multicolumn{4}{l}{
\textit{Proposed method and variant}
} \\
PLatour-SFT &
Sender hidden states &
Projectors &
SFT \\

\textbf{PLatour} &
\textbf{Sender hidden states} &
\textbf{Projectors} &
\textbf{GRPO} \\

\bottomrule
\end{tabularx}
\caption{
Comparison of baseline methods and PLatour. The table distinguishes the source of information, trainable components, and optimization objectives. }
\label{tab:baselines}
\end{table}

\subsection{Training Settings}
\label{app:training}

We use AdamW with $\beta=(0.9,0.999)$ and gradient-norm clipping at 1. Table~\ref{tab:training_settings} summarizes the optimization settings for the Qwen3 experiments. Full-parameter variants use a learning rate ten times lower than the corresponding LoRA or bridge-only setting. They update every receiver weight---more than an order of magnitude more trainable parameters than LoRA at rank 128 or PLatour's per-layer projections---and are correspondingly more likely to destabilize the pretrained model; the lower rate follows standard practice for full-parameter fine-tuning.

\begin{table}[H]
    \centering
    \small
    \begin{tabular}{lccc}
        \toprule
        Setting & SFT & KV-Lingo & GRPO \\
        \midrule
        Epochs        & 3 & 3 & 3 \\
        Global batch size & 64 & 64 & 64 / 32 \\
        Learning rate & $10^{-4}$ & $10^{-4}$ & $10^{-5}$ \\
        LR schedule   & Linear & Cosine & Constant \\
        Warmup ratio  & 10\% & 5\% & 0 \\
        \bottomrule
    \end{tabular}
    \caption{Training hyperparameters. GRPO batch sizes count questions, each with eight sampled responses; Chemistry uses 32 questions per batch.}
    \label{tab:training_settings}
\end{table}

\paragraph{Supervised training.}
PLatour, C2C, and LoRA use response-only cross-entropy, excluding prompt and padding tokens. Weight decay is zero for PLatour and 0.01 for C2C and LoRA. LoRA uses rank 128, scaling parameter 128, and dropout 0.1. We evaluate the final third-epoch checkpoint.

\paragraph{KV-Lingo.}
We initialize the translators using closed-form fitting on 500 training prompts, followed by three epochs of full-vocabulary KL refinement with zero weight decay.

\paragraph{GRPO.}
We use the verl framework~\citep{verl} and sample eight responses per question. Advantages are normalized within each group using the sample standard deviation and $\epsilon=10^{-6}$. Each rollout batch receives one policy optimization epoch, with clipping parameter 0.2, dual clipping at 3, and token-level rollout importance weights capped at 2. Weight decay, KL regularization, and entropy bonuses are disabled. Training rollouts use temperature 1 without top-$p$ or top-$k$ truncation. For Qwen3, we train on the complete source training pools, with response limits of 2,048 tokens for Chemistry and GSM8K and 4,096 for MATH. LoRA-GRPO uses rank and scaling parameter 128, with dropout disabled. For Qwen3.5, we train for one complete epoch: 264 optimizer steps on DAPO and 295 on DeepCoder. Both tasks use a question batch size of 64, a learning rate of $10^{-5}$, and a response limit of 16,384 tokens. Full-parameter GRPO uses $10^{-6}$; every other Qwen3.5 configuration, including the bridge, LoRA, self-communication, and virtual-sender variants, uses $10^{-5}$. DAPO rewards require a correct answer on the final non-empty line in the form \texttt{Answer: X}; DeepCoder rewards require all test cases to pass. Length-truncated responses receive zero training reward. Checkpoint selection follows the criterion described in Section~\ref{app:evaluation}.

\subsection{Evaluation Protocol}
\label{app:evaluation}

All evaluations disable thinking mode. For each task, adapted models and unadapted baselines use the same prompts, decoding settings, sample counts, and response budgets, summarized in Table~\ref{tab:evaluation_settings}. Qwen3 sampled decoding uses temperature 0.7, \texttt{top\_p}=0.8, \texttt{top\_k}=20, and zero presence penalty. Qwen3.5 uses temperature 1, \texttt{top\_p}=0.95, \texttt{top\_k}=20, and presence penalty 1.5. Both use \texttt{min\_p}=0, zero frequency penalty, and repetition penalty 1. For SFT, we report the last checkpoint's test score. For each GRPO configuration, we select the evaluated checkpoint with the highest mean validation accuracy, and use it for test set evaluation.

\begin{table}[htbp]
    \centering
    \small
    \begin{tabular}{llrr}
        \toprule
        Task & Decoding & $k$ & Token limit \\
        \midrule
        \multicolumn{4}{l}{\textit{Qwen3}} \\
        ARC-C (val/test)             & Greedy  & 1 & 2,048 \\
        MMLU (val/test)               & Greedy  & 1 & 64    \\
        Chemistry (no-CoT) (val/test)  & Greedy  & 1 & 64    \\
        Chemistry (CoT) (val/test)    & Sampled & 8 & 2,048 \\
        GSM8K (val/test)              & Sampled & 2 & 2,048 \\
        MATH500 (val/test)            & Sampled & 4 & 4,096 \\
        \midrule
        \multicolumn{4}{l}{\textit{Qwen3.5}} \\
        DAPO (val)        & Sampled & 4  & 16,384 \\
        DeepCoder (val)   & Sampled & 16 & 16,384 \\
        AIME (135 / 80)    & Sampled & 16 & 32,768 \\
        HMMT 2025          & Sampled & 16 & 32,768 \\
        IMO-AnswerBench v2 & Sampled & 4  & 32,768 \\
        LiveCodeBench v6 increment
                           & Sampled & 8  & 16,384 \\
        HumanEval          & Sampled & 8  & 16,384 \\
        MBPP               & Sampled & 8  & 16,384 \\
        \bottomrule
    \end{tabular}
        \caption{Evaluation settings. $k$ denotes responses per question; token limits apply to individual responses. Rows marked ``val'' denote evaluations on the validation partitions; reported GRPO checkpoints are selected by mean validation accuracy. AIME includes both the full 135-problem set and the 80-problem subset excluding DAPO overlaps.}
    \label{tab:evaluation_settings}
\end{table}

\paragraph{Answer scoring.}
For ARC-C, MMLU, and Chemistry no-CoT, we extract the boxed answer option and compare it with the gold label. Chemistry CoT follows SDPO's final \texttt{<answer>} field extraction. GSM8K and MATH500 use the final complete boxed answer, with mathematical equivalence checked using \texttt{math-verify}. For Qwen3.5 mathematical evaluation, we extract the last \texttt{Answer:} field and verify the complete answer expression using the local \texttt{math-verify} scorer, including on IMO-AnswerBench. Code responses are executed against the benchmark test cases and are counted as correct only if all cases pass. Missing or invalid answers are scored as incorrect. Length-truncated responses are retained in all evaluations. They follow the standard scoring rules for Qwen3 and code benchmarks, but are scored as incorrect for Qwen3.5 mathematical evaluation. The training-only truncation penalty is not applied to code evaluation.

\paragraph{Metrics.}
For $N$ questions with $k$ responses each, let $c_{ij}\in\{0,1\}$ indicate whether response $j$ to question $i$ is correct. We report
\begin{equation}
    \operatorname{avg}@k
    = \frac{1}{Nk}
      \sum_{i=1}^{N}\sum_{j=1}^{k} c_{ij},
\end{equation}
which estimates single-response accuracy rather than best-of-$k$ success. For greedy decoding, this reduces to standard accuracy. Where reported, we additionally use
\begin{equation}
    \operatorname{pass}@k
    = \frac{1}{N}\sum_{i=1}^{N}
      \mathbf{1}\!\left[
          \max_{1\leq j\leq k} c_{ij}=1
      \right].
\end{equation}
Sampled evaluations use fixed per-question and per-sample seeds, shared across compared methods within each benchmark.

\subsection{AIME Year and HMMT Competition Breakdowns}
\label{app:math_breakdowns}

AIME-135 combines AIME I and II from 2022--2025 (30 problems per year) and AIME I from 2026 (15 problems). Table~\ref{tab:aime_year_full} reports each year's \(\operatorname{avg}@16\) and the response-weighted full-set result. The fixed AIME-80 subset excludes 53 problems overlapping with DAPO training and two overlapping with validation. Its remaining question counts by year are 2, 4, 30, 30, and 14, respectively. Table~\ref{tab:aime_year_clean} shows this same predefined subset by year; the 2022 and 2023 clean-year columns have very small denominators and should not be interpreted as representative yearly benchmarks. HMMT 2025 combines February and November, with 30 problems per competition; Table~\ref{tab:hmmt_competition} reports each separately. The year-wise and competition-wise columns decompose the results of the same checkpoints reported in Table~\ref{tab:qwen35_res}, using the same saved responses and evaluation rules. No separate checkpoint is selected for an individual year or competition. IMO-AnswerBench v2 is one benchmark, and the three coding benchmarks are already reported individually in Table~\ref{tab:qwen35_res}.

\begin{table*}[p]
    \centering
    \small
    \setlength{\tabcolsep}{4pt}
    \renewcommand{\arraystretch}{1.15}
    \begin{tabular}{@{}lrrrrrr@{}}
        \toprule
        Method & 2022 & 2023 & 2024 & 2025 & 2026 I & AIME-135 \\
        \midrule
        Receiver-Alone & 60.00 & 57.29 & 67.92 & 52.50 & 56.67 & 59.12 \\
        PLatour (397B) & 73.54 & 73.33 & 77.92 & 65.63 & 73.75 & 72.73 \\
        PLatour (35B) & 73.13 & 72.08 & 81.25 & 70.21 & 78.33 & 74.63 \\
        Full-GRPO & 75.83 & 76.88 & 82.08 & 72.08 & 79.58 & 77.04 \\
        LoRA-GRPO & 74.38 & 73.75 & 78.96 & 65.21 & 77.50 & 73.56 \\
        Self-Communication& 73.13 & 70.21 & 75.83 & 68.33 & 76.67 & 72.41 \\
        Virtual Sender & 72.71 & 71.67 & 77.50 & 67.50 & 75.83 & 72.73 \\
        \textit{Sender-Alone (35B)} & 78.54 & 73.33 & 85.83 & 75.42 & 82.08 & 78.70 \\
        \textit{Sender-Alone (397B)} & 87.92 & 86.04 & 88.33 & 84.58 & 83.75 & 86.39 \\
        \bottomrule
    \end{tabular}
    \caption{AIME-135 accuracy (\%) by year, measured as \(\operatorname{avg}@16\). Each of 2022--2025 contains 30 questions (480 responses), while 2026 contains 15 AIME I questions (240 responses). The full-set column pools all 135 questions (2,160 responses) and is not the unweighted mean of the five displayed yearly percentages. All rows use the 32K response limit and the original Qwen3.5 mathematical answer scorer.}
    \label{tab:aime_year_full}
\end{table*}

\begin{table*}[p]
    \centering
    \small
    \setlength{\tabcolsep}{4pt}
    \renewcommand{\arraystretch}{1.15}
    \begin{tabular}{@{}lrrrrrr@{}}
        \toprule
        Method & 2022 & 2023 & 2024 & 2025 & 2026 I & AIME-80 \\
        \midrule
        Receiver-Alone & 68.75 & 59.38 & 67.92 & 52.50 & 53.57 & 59.22 \\
        PLatour (397B) & 78.13 & 70.31 & 77.92 & 65.63 & 71.88 & 71.88 \\
        PLatour (35B) & 71.88 & 71.88 & 81.25 & 70.21 & 76.79 & 75.63 \\
        Full-GRPO & 68.75 & 75.00 & 82.08 & 72.08 & 78.13 & 76.95 \\
        LoRA-GRPO & 62.50 & 76.56 & 78.96 & 65.21 & 75.89 & 72.73 \\
        Self-Communication& 62.50 & 73.44 & 75.83 & 68.33 & 75.00 & 72.42 \\
        Virtual Sender & 75.00 & 68.75 & 77.50 & 67.50 & 74.11 & 72.66 \\
        \textit{Sender-Alone (35B)} & 75.00 & 73.44 & 85.83 & 75.42 & 80.80 & 80.16 \\
        \textit{Sender-Alone (397B)} & 93.75 & 84.38 & 88.33 & 84.58 & 82.59 & 85.86 \\
        \bottomrule
    \end{tabular}
    \caption{AIME-80 accuracy (\%) by year on the fixed, decontaminated subset of AIME-135. The retained question counts for 2022--2026 are 2, 4, 30, 30, and 14 (32, 64, 480, 480, and 224 responses at \(k=16\)). The last column pools 80 questions (1,280 responses); it is not an equal-weight average of years. In particular, the 2022 and 2023 entries describe only two and four questions, not their complete annual AIME competitions. Scores retain the original 32K generation and answer-equivalence protocol.}
    \label{tab:aime_year_clean}
\end{table*}

\begin{table*}[p]
    \centering
    \small
    \setlength{\tabcolsep}{5pt}
    \renewcommand{\arraystretch}{1.15}
    \begin{tabular}{@{}lrrrrrr@{}}
        \toprule
        & \multicolumn{3}{c}{\(\operatorname{avg}@16\) (\%)}
        & \multicolumn{3}{c}{\(\operatorname{pass}@16\) (\%)} \\
        \cmidrule(lr){2-4}\cmidrule(l){5-7}
        Method & Feb. & Nov. & All 60 & Feb. & Nov. & All 60 \\
        \midrule
        Receiver-Alone & 39.79 & 52.92 & 46.35 & 80.00 & 83.33 & 81.67 \\
        PLatour (397B) & 54.38 & 69.17 & 61.77 & 86.67 & 90.00 & 88.33 \\
        PLatour (35B) & 58.75 & 70.42 & 64.58 & 90.00 & 90.00 & 90.00 \\
        Full-GRPO & 60.63 & 69.79 & 65.21 & 90.00 & 86.67 & 88.33 \\
        LoRA-GRPO& 56.46 & 62.50 & 59.48 & 90.00 & 86.67 & 88.33 \\
        Self-Communication  & 50.42 & 63.54 & 56.98 & 83.33 & 90.00 & 86.67 \\
        Virtual Sender & 53.75 & 65.83 & 59.79 & 76.67 & 90.00 & 83.33 \\
        \textit{Sender-Alone (35B)} & 65.00 & 74.38 & 69.69 & 86.67 & 93.33 & 90.00 \\
        \textit{Sender-Alone (397B)} & 77.50 & 87.08 & 82.29 & 93.33 & 93.33 & 93.33 \\
        \bottomrule
    \end{tabular}
    \caption{HMMT 2025 results by competition, using the same 30 February and 30 November problems for every method. Each competition has 480 responses, so \(\operatorname{avg}@16\) is the fraction of correct responses and \(\operatorname{pass}@16\) is the fraction of questions answered correctly at least once; ``All 60'' pools both competitions rather than treating them as separate benchmarks in the three-test math average. Sender-Alone 397B reaches 82.29\% \(\operatorname{avg}@16\) and 93.33\% \(\operatorname{pass}@16\) locally; the two metrics must not be conflated. For Receiver-Alone and Sender-Alone 35B, only responses previously truncated at 16K were independently regenerated at 32K; their other saved responses were reused. All other rows use the original full 32K test runs.}
    \label{tab:hmmt_competition}
\end{table*}
\clearpage

\subsection{Prompt Templates}
\label{app:prompts}

All prompts use the models' native chat templates with thinking mode disabled and no in-context demonstrations. The templates below show user messages unless otherwise specified. Braced placeholders denote task-specific content.

\lstdefinestyle{taskprompt}{
    language={},
    basicstyle=\small\ttfamily,
    numbers=none,
    frame=tb,
    breaklines=true,
    breakatwhitespace=true,
    columns=fullflexible,
    keepspaces=true,
    showstringspaces=false,
    mathescape=false,
    keywordstyle={},
    commentstyle={},
    stringstyle={}
}

\paragraph{Qwen3: ARC-C, MMLU, and Chemistry (no-CoT).}
Training and evaluation use the same answer-only template. Option rows are included only when present.

\begin{lstlisting}[style=taskprompt]
{Question}

A. {Option A}
B. {Option B}
...

Reply with exactly "The answer is \boxed{X}", replacing X with the letter of the correct option. Do not include any explanation or any other text.
\end{lstlisting}

\paragraph{Qwen3: Chemistry (CoT).}
We preserve the SDPO system message~\citep{sdpo}, which requires reasoning inside \texttt{<reasoning>...</reasoning>} and the final option letter inside \texttt{<answer>...</answer>}. Trajectory generation, GRPO training, and evaluation use the following user-message format:

\begin{lstlisting}[style=taskprompt]
{Question}

A: {Option A}
B: {Option B}
...

Please reason step by step.
\end{lstlisting}

\paragraph{Qwen3: GSM8K and MATH.}
The following template is shared by supervised trajectory generation, GRPO training, and evaluation, including MATH500.

\begin{lstlisting}[style=taskprompt]
{Question}

Please reason step by step, and put your final answer within \boxed{}.
\end{lstlisting}

\paragraph{Qwen3.5: DAPO and mathematical benchmarks.}
We retain the original DAPO instruction for training and validation. AIME, HMMT, and IMO-AnswerBench evaluations use the same training-aligned format:

\begin{lstlisting}[style=taskprompt]
Solve the following math problem step by step. The last line of your response should be of the form Answer: $Answer (without quotes) where $Answer is the answer to the problem.

{Question}

Remember to put your answer on its own line after "Answer:".
\end{lstlisting}

\paragraph{Qwen3.5: DeepCoder and LiveCodeBench.}
DeepCoder training and validation and LiveCodeBench evaluation use the official LiveCodeBench generic system message and user-message format~\citep{lcb}. The format instruction depends on whether starter code is provided; both the instruction and system message are retained from the pinned LiveCodeBench implementation.

\begin{lstlisting}[style=taskprompt]
### Question:
{Problem statement}

### Format: {Official format instruction}
```python
{Starter code, or # YOUR CODE HERE}
```

### Answer: (use the provided format with backticks)
\end{lstlisting}

\paragraph{Qwen3.5: HumanEval and MBPP.}
HumanEval evaluation provides the original function prompt, including its signature and docstring:

\begin{lstlisting}[style=taskprompt]
Please provide a self-contained Python script that solves the following problem in a markdown code block:
```python
{Original HumanEval function prompt}
```
\end{lstlisting}

MBPP evaluation provides the problem statement and its three supplied test assertions:

\begin{lstlisting}[style=taskprompt]
Please provide a self-contained Python script that solves the following problem in a markdown code block:
```python
"""
{Problem statement}
{Assertion 1}
{Assertion 2}
{Assertion 3}
"""
```
\end{lstlisting}

\subsection{Dummy Transfer and Irrelevant Transfer Settings}
\label{app:sender_controls}

We evaluate both controls using the original trained checkpoint, without additional training. All model and communication parameters remain frozen. We preserve the original layer mapping, prompts, decoding settings, response budgets, and scoring rules. The receiver always receives the original test problem, and its outputs are scored against that problem's reference answer or executable tests.

\paragraph{Dummy transfer.}
We replace question-dependent sender features with fixed template features computed offline. We select 500 distinct prompts from the checkpoint's training pool using seed 42, excluding validation and test examples. Using prompt-only sender forwards, we compute elementwise mean features separately for each communicated layer and each fixed prefix/suffix token position. Suffix positions are aligned relative to the template suffix rather than by absolute sequence position.

At inference, no sender model is loaded or called. The original communication modules transfer these fixed means only at template positions. Direct communication is disabled at question positions, padding, the final prompt token, and response positions. For PLatour, the means are computed from sender block-input hidden states before RMSNorm. For KV-based methods, keys are averaged before RoPE: C2C uses normalized keys, whereas KV-Lingo uses raw keys; both also average values. The original method-specific transformations and positional encoding are applied at the actual test positions.

\paragraph{Irrelevant transfer.}
For each test problem, we select an unrelated training problem with the same fully rendered prompt-token length, including all instructions and chat-template tokens. Identical problems and known duplicates or paraphrases are excluded. Donors are selected deterministically using seed 42, with replacement, and the assignment is fixed across samples and checkpoints sharing the same protocol. When no exact-length donor is available, we rephrase only the donor's wording to match the required length, preserving its problem semantics and requirements. Rewriting uses only the donor and target length, without access to the test problem or its answer.

The sender processes the donor prompt, while the receiver processes the original test prompt. Donor features are transferred at all positions eligible under original transfer, including question positions. No donor solution is generated or transferred. For both controls, communication excludes the final prompt token and occurs only during prefill; subsequent decoding uses the resulting receiver cache.

\section{TextComm Prompt Selection}
\label{app:textcomm}

TextComm transmits text to the receiver instead of hidden states. The sender reads the problem and writes a message, and the receiver generates the response from the problem followed by this message. Both models are frozen and run in non-thinking mode. The text baseline of C2C~\citep{c2c}, T2T, asks the sender for one sentence of background knowledge. Before fixing the TextComm prompt, we compared T2T with four other prompts. No prompt allows the sender to solve the problem.

\paragraph{Prompts.}
The five prompts differ only in the sender instruction.
\begin{itemize}
  \item \textbf{T2T}~\citep{c2c}: the original C2C prompt, which asks for one sentence of the most essential background knowledge.
  \item \textbf{Step-Back}~\citep{stepback}: the general concepts, principles, and solution techniques required by the problem, in at most 120 words and without the specific numbers in the problem.
  \item \textbf{Plan}~\citep{planandsolve}: a plan of at most five steps that states what to compute in each step but not the results.
  \item \textbf{Hint}~\citep{hsp}: the key idea, a relevant formula or theorem, or a useful first step.
  \item \textbf{Understand}: a restatement of what is asked, the given conditions and implicit constraints, the relevant concepts, and likely pitfalls, in at most 150 words.
\end{itemize}
Except for T2T, each prompt instructs the sender not to solve the problem, not to carry out calculations, and not to state or imply the final answer or any intermediate numerical result. The receiver prompt is:
\begin{quote}\small\raggedright
{\ttfamily \{problem\}\\
A more capable assistant read this problem and left the following note (it may be incomplete and is not a full solution):\\
<note> \{message\} </note>}\\
followed by the answer-format instruction of the benchmark.
\end{quote}

\paragraph{Experimental Setup.}
The receiver is Qwen3.5-4B. We use two senders: Qwen3.5-35B-A3B, from the receiver's model family, and GLM-5.3-Flash, from a different family. The Qwen3.5-35B-A3B sender runs in non-thinking mode with $T{=}0.7$, top-$p{=}0.8$, top-$k{=}20$, and presence penalty $1.5$. \textbf{We access GLM-5.3-Flash through OpenRouter\footnote{\url{https://openrouter.ai/}}, where its reasoning cannot be disabled}. We therefore use the lowest reasoning effort with $T{=}1.0$ and top-$p{=}0.95$. Its hidden reasoning averages 40--250 tokens per message across prompts and benchmarks and is not passed to the receiver. Each problem receives one message, which is shared by all receiver samples. The receiver uses the non-thinking sampling settings recommended by Qwen and the same per-sample seeds as Receiver-Alone. We select the prompt on MATH500 (avg@4, 4K-token limit) and AIME 2022--2026 (avg@16, 16K-token limit), with answers in \verb|\boxed{}|. Table~\ref{tab:textcomm-selection} reports the results. Gains are measured against Receiver-Alone on the same hardware, with 95\% confidence intervals from a paired bootstrap over problems (5,000 resamples).

\begin{table}[htbp]
\centering
\footnotesize
\setlength{\tabcolsep}{5pt}
\begin{tabular}{@{}llrrr@{}}
\toprule
Sender & Prompt & MATH500 & AIME & Tok. \\
\midrule
-- & Receiver-Alone & 85.65 & 58.61 & -- \\
\midrule
\multirow{5}{*}{Qwen3.5-35B-A3B}
 & T2T        & 86.10 (+0.45 [$-$0.85, 1.75]) & 58.98 (+0.37 [$-$1.81, 2.55]) & 47--58 \\
 & Step-Back  & 85.75 (+0.10 [$-$1.50, 1.65]) & --                            & 138 \\
 & Plan       & 86.15 (+0.50 [$-$1.20, 2.20]) & --                            & 134 \\
 & Hint       & 86.05 (+0.40 [$-$1.25, 2.05]) & 61.48 (+2.87 [$-$0.09, 5.79]) & 102--265 \\
 & Understand & \textbf{87.30} (+1.65 [0.25, 3.05]) & \textbf{61.76} (+3.15 [0.60, 5.74]) & 274--299 \\
\midrule
\multirow{3}{*}{GLM-5.3-Flash}
 & T2T        & 86.40 (+0.75 [$-$0.75, 2.35]) & 59.63 (+1.02 [$-$1.30, 3.33]) & 49--70 \\
 & Hint       & 87.15 (+1.50 [$-$0.15, 3.15]) & 60.60 (+1.99 [$-$1.02, 4.81]) & 71--129 \\
 & Understand & \textbf{88.00} (+2.35 [0.85, 3.80]) & \textbf{63.84} (+5.23 [3.10, 7.50]) & 293--334 \\
\bottomrule
\end{tabular}
\caption{TextComm prompt selection via Qwen3.5-4B receiver. Each cell gives avg@$k$, followed by the gain over Receiver-Alone and its 95\% confidence interval. Tok.\ is the average number of message tokens seen by the receiver (MATH500--AIME). Step-Back and Plan are evaluated only on MATH500 with the Qwen3.5-35B-A3B sender.}
\label{tab:textcomm-selection}
\end{table}

\paragraph{Results.}
T2T barely changes receiver accuracy. Its gains are below 1.1 points for both senders on both benchmarks, and all confidence intervals include zero. Step-Back, Plan, and Hint also do not improve the receiver consistently. Understand gives the largest gain in every setting and is the only prompt whose gain is significant for both senders on both benchmarks. Its gain is larger on AIME than on MATH500 and larger with the GLM-5.3-Flash sender. Longer messages tend to give larger gains, but the Understand message (about 300 tokens) remains short compared with a receiver response (about 1.5K tokens on MATH500 and 7.7K tokens on AIME).

\paragraph{Answer leakage.}
A stronger sender may reveal the answer in its message. We flag a message if it contains a \verb|\boxed{}| expression, a statement of the form ``the answer is $\langle$value$\rangle$'', or the normalized gold answer when the answer does not appear in the problem. We then inspect every flagged message manually. Most flags are false positives: the gold value is restated from the problem, appears inside a standard formula such as the reflection $(x,y)\mapsto(x,2k-y)$, or is given as an incorrect value to avoid. We find three leaked answers, all from Understand with the Qwen3.5-35B-A3B sender on MATH500. Excluding these three problems, Understand scores 87.22 and Receiver-Alone scores 85.61.

\paragraph{Choice of prompt.}
We use Understand for TextComm. Unlike PLatour, TextComm requires the sender to decode a message (about 300 tokens for Understand). The comparison is therefore limited to short messages; it does not cover prompting methods in which the sender first reasons at length.

\section{TextComm Results}
\label{app:textcomm-results}

All TextComm results below use the Understand prompt. %These runs use H100 and H200 GPUs, while the other results in Tables~\ref{tab:qwen3_res} and~\ref{tab:qwen35_res} were obtained on B200 GPUs, so we also report Receiver-Alone on our hardware and measure TextComm gains against it.

\paragraph{Qwen3.5 mathematical reasoning.}
TextComm follows the evaluation setting of Table~\ref{tab:qwen35_res}: the \texttt{Answer:} format, a 32K-token limit, and the same scorer and seeds. The note is inserted between the problem and the final answer-format reminder; without a note, the prompt is token-identical to the original. The Qwen3.5-35B-A3B and Qwen3.5-397B-A17B senders are accessed through OpenRouter with reasoning disabled, using the sender sampling settings above. The 4B sender is the receiver model itself. Table~\ref{tab:textcomm-qwen35} reports the results together with T2T for comparison with C2C.

\begin{table}[htbp]
\centering
\footnotesize
\setlength{\tabcolsep}{6pt}
\begin{tabular}{@{}llrrr@{}}
\toprule
Sender & Prompt & AIME & HMMT & IMO-Ans. \\
\midrule
-- & Receiver-Alone & 61.25 & 44.27 & 38.63 \\
\midrule
\multirow{2}{*}{397B}
 & T2T        & 60.46 ($-$0.79) & 46.56 (+2.29) & 41.69 (+3.06) \\
 & Understand & 63.15 (+1.90) & 49.06 (+4.79) & 42.06 (+3.44) \\
\midrule
\multirow{2}{*}{35B}
 & T2T        & 61.30 (+0.05) & 45.63 (+1.35) & 39.88 (+1.25) \\
 & Understand & 62.92 (+1.67) & 46.77 (+2.50) & 41.94 (+3.31) \\
\midrule
4B & Understand & 57.36 ($-$3.89) & 43.96 ($-$0.31) & 39.63 (+1.00) \\
\bottomrule
\end{tabular}
\caption{TextComm on the Qwen3.5 math benchmarks via Qwen3.5-4B receiver (\texttt{Answer:} format, 32K tokens). Gains over Receiver-Alone on the same hardware are in parentheses. IMO-Ans.\ denotes IMO-AnswerBench v2.}
\label{tab:textcomm-qwen35}
\end{table}

Understand improves the receiver by 1.7 to 4.8 points with the 35B and 397B senders, and T2T by $-$0.8 to 3.1 points. On AIME, neither gain is significant (95\% confidence intervals include zero), whereas PLatour with the same senders improves the receiver by 15.51 (35B) and 13.61 (397B) points on AIME-135, and by 16.41 and 12.66 points on the decontaminated AIME-80 (Table~\ref{tab:qwen35_res}). When the 4B receiver writes the message for itself, AIME drops by 3.89 points (95\% CI [$-$6.94, $-$0.93]), so the gains with larger senders come from their messages rather than from restating the problem. The screen in Appendix~\ref{app:textcomm} flags no message on AIME or HMMT. On IMO-AnswerBench, one message cites Euler's formula $CI^2=R(R-2r)$, which contains the answer expression $R-2r$. We count it as a corner case.

\paragraph{Qwen3 benchmarks.}
TextComm follows the evaluation setting of each column in Table~\ref{tab:qwen3_cross_model_transfer}, including prompts, decoding, the number of samples, and the scorer. For ARC-C, MMLU-Redux, and Chemistry, the sender sees the question and its options and uses a task-agnostic wording of the Understand prompt, which also forbids indicating or ruling out options; GSM8K and MATH500 use the prompt above. The note is inserted between the question (with its options) and the answer-format instruction. The 8B and 1.7B senders are Qwen3-8B and Qwen3-1.7B in non-thinking mode. The original Qwen3-235B-A22B is no longer served, so the 235B sender is Qwen3-235B-A22B-Instruct-2507, accessed through OpenRouter. On our hardware, Receiver-Alone scores 74.32 (ARC-C), 53.92 (MMLU-Redux), 39.05 (Chemistry without CoT), 39.11 (Chemistry with CoT), 82.37 (GSM8K), and 72.90 (MATH500), which match Table~\ref{tab:qwen3_cross_model_transfer} within 0.1 points on ARC-C, MMLU-Redux, and Chemistry without CoT, and within 2.3 points elsewhere.

Relative to Receiver-Alone on the same hardware, TextComm improves ARC-C by 5.5 to 7.5 points and MMLU-Redux by 4.4 to 6.3 points, including with the 1.7B receiver writing to itself. It does not improve Chemistry without CoT: with a note in its context, the receiver more often adds an explanation and reaches the 64-token limit. On Chemistry with CoT, it improves the receiver by 4.2 points with the 8B sender and by 1.7 points with the 235B sender. On GSM8K, the 235B and 8B senders add 7.9 and 5.8 points, respectively, and on MATH500, 3.2 and 1.5 points.

\paragraph{Answer leakage on the Qwen3 benchmarks.}
On GSM8K, the senders often carry out arithmetic despite the instruction. The gold answer appears in 16 (235B), 20 (8B), and 80 (1.7B) of the 1,319 messages without appearing in the problem. Excluding these problems changes the gains by at most 0.4 points. On ARC-C, MMLU-Redux, and Chemistry, the screen flags at most 0.5\% of the messages from the 235B and 8B senders. The 1.7B sender is flagged on 3.5\% (ARC-C) and 3.4\% (MMLU-Redux) of the questions. Excluding flagged questions changes any gain on these benchmarks by at most 0.3 points.

\section{Design Evolution of PLatour: Ablation Studies}
\label{sec:appendix}
% bring the ablation sec to here, and make it as the progress of the dev.
In this section, we describe the design trajectory from C2C \citep{c2c} to PLatour, which extends PACT from improving only multiple-choice question answering to performing well on long-form generation tasks such as MATH, AIME, and coding, while also supporting transfer to linear-attention layers. We conduct all ablation studies on MATH, using Qwen3-30B and Qwen3.5-122B as senders and Qwen3-4B and Qwen3.5-4B as receivers.

\begin{table}[htbp]
\centering
\footnotesize
\setlength{\tabcolsep}{8pt}
\begin{tabular}{@{}lllc@{}}
\toprule
Standalone baselines
& \multicolumn{3}{l@{}}{
    Qwen3-4B: 84.60 \qquad Qwen3-30B: 88.65 \quad Qwen3.5-4B: 83.30 \qquad Qwen3.5-122B: 93.15
} \\
\midrule
Ablation & Variant & Sender $\rightarrow$ Receiver
& MATH500 \\

\midrule

Hidden states or KV caches
& \textbf{Hidden-state transfer} & Qwen3-30B $\rightarrow$ Qwen3-4B & \textbf{88.05} \\
& Hidden-state transfer (no propagate) & Qwen3-30B $\rightarrow$ Qwen3-4B & 86.80\\
& KV-cache transfer & Qwen3-30B $\rightarrow$ Qwen3-4B & 86.00 \\
& KV-cache transfer (propagate) & Qwen3-30B $\rightarrow$ Qwen3-4B & 85.05 \\
& Original C2C recipe + on-policy
& Qwen3-30B $\rightarrow$ Qwen3-4B & 86.85 \\
& Original C2C recipe + on-policy (soft)
& Qwen3-30B $\rightarrow$ Qwen3-4B & 87.25 \\
\midrule

Reward maximization
& \textbf{GRPO} & Qwen3-30B $\rightarrow$ Qwen3-4B & \textbf{88.05} \\
or distillation
& SFT & Qwen3-30B $\rightarrow$ Qwen3-4B & 83.55 \\
& OPD & Qwen3-30B $\rightarrow$ Qwen3-4B & 83.20 \\
\midrule
Layer matching strategies
& \textbf{Relative-depth layers (full-to-full)} & Qwen3.5-122B $\rightarrow$ Qwen3.5-4B & \textbf{91.25} \\
& Relative-depth layers
& Qwen3.5-122B $\rightarrow$ Qwen3.5-4B & 89.70 \\
& Relative-depth blocks
& Qwen3.5-122B $\rightarrow$ Qwen3.5-4B & 88.00 \\
& Two-layer mapping (full-to-full)
& Qwen3.5-122B $\rightarrow$ Qwen3.5-4B & 89.85 \\
& Last-N layers
& Qwen3.5-122B $\rightarrow$ Qwen3.5-4B & 90.75 \\
\midrule
Projection network
& \textbf{Linear} & Qwen3-30B $\rightarrow$ Qwen3-4B & \textbf{88.05} \\
& Two-layer Linear & Qwen3-30B $\rightarrow$ Qwen3-4B & 85.40 \\
& Two-layer MLP (GELU)& Qwen3-30B $\rightarrow$ Qwen3-4B & 85.80 \\
\midrule
Projection initialization
& \textbf{Zero initialization} & Qwen3-30B $\rightarrow$ Qwen3-4B & \textbf{88.05} \\
& Random initialization & Qwen3-30B $\rightarrow$ Qwen3-4B & 85.05  \\
\bottomrule
\end{tabular}
\caption{
Ablation studies with training on the MATH dataset and evaluation on MATH500, reporting avg@4. Within each ablation group, all other experimental settings are held fixed. PLatour's final configurations are bolded. }
\label{tab:ablation}
\end{table}

\paragraph{Hidden States or KV Caches.} 
We first replace key--value (KV) caches with hidden states to make the method compatible with linear-attention layers, which allows us to run experiments with Qwen3.5 receivers in line with Mostik's setting. As shown in Table~\ref{tab:ablation}, hidden-state transfer even surpasses KV-cache transfer in all combinations we tried.

\paragraph{Reward Maximization or Distillation?}
This is another major departure from previous methods: previous SFT-based methods degrade generation performance relative to the raw receiver, as shown in Table~\ref{tab:qwen3_cross_model_transfer}. Our initial SFT variant of PLatour surpasses the raw receiver on two short-generation tasks (Chemistry and GSM8K), but still degrades performance on long-generation tasks such as MATH (Tables~\ref{tab:ablation} and~\ref{tab:qwen3_cross_model_transfer}). Only GRPO consistently improves over the raw receiver by a large margin. Moreover, OPD also degrades performance. These results suggest that distillation fails in the PACT setting. Our interpretation is that it is difficult for the small receiver to imitate the large sender's output distribution, whereas it is easier for the receiver to leverage useful information from the sender as a `tool' to obtain higher rewards. This key design choice enables PLatour to improve over the raw receiver on long-form generation tasks.

\paragraph{Other Micro-Designs.} We find that relative-depth layer match performs the best, and requiring full-attention receiver layers to match full-attention sender layers improves performance. However, for the supervised fine-tuning (SFT) variant (PLatour-SFT), C2C's layer matching strategy \citep{c2c}, pairing the receiver's $L$ layers with the final $L$ sender layers, works better in SFT but not GRPO. Thus, we adopt them in all SFT training. In addition, a linear projection layer performs better than a two-layer linear projector and a two-layer MLP. We think a linear projection may have enough expressiveness when pre-trained models share a similar representation space \citep{plato,oprd}, and are easier for optimization especially in the direct GRPO training. Moreover, we find that initialization plays an important role here because zero-initialization makes the initial receiver's behavior the same. More details are discussed in Appendix \ref{app:ablation}.

\FloatBarrier
\section{Additional Ablations and Analysis}
\label{app:ablation}

\subsection{More on Layer Matching}
\label{app:layer_matching}

We compare two layer-matching strategies in more detail: relative-depth matching distributes receiver layers across the sender's depth, whereas last-$n$ matching aligns the final $n$ sender layers with the $n$ receiver layers.

As shown in Table~\ref{tab:layer_matching_ablation}, last-$n$ matching improves PLatour-SFT performance on ARC-C, MMLU-Redux, and GSM8K for both sender sizes. The Chemistry results are mixed: last-$n$ matching improves CoT performance with the 235B sender but reduces both no-CoT and CoT performance with the 8B sender. For the 235B-to-1.7B GSM8K GRPO experiment, relative-depth matching achieves 89.35\%, exceeding last-$n$ matching by 2.85 percentage points. Moreover, in the Qwen3.5-122B $\rightarrow$ Qwen3.5-4B setting, Table~\ref{tab:ablation} reports similar accuracies for the two strategies (90.75 and 91.25), while Table~\ref{tab:grpo_layer_match} shows that relative-depth (full-to-full) matching reaches high accuracy earlier in training. These results suggest that the effectiveness of layer matching depends on both the task and the training objective.

\begin{table}[htbp]
    \centering
    \footnotesize
    \setlength{\tabcolsep}{6pt}
    \begin{tabular}{@{}llrrrrr@{}}
        \toprule
        & & \multicolumn{3}{c}{Answer-only}
          & \multicolumn{2}{c}{CoT Generation} \\
        \cmidrule(lr){3-5}
        \cmidrule(l){6-7}
        Method & Layer Matching
        & ARC-C & MMLU-Redux & Chemistry
        & Chemistry & GSM8K \\
        \midrule
        \multicolumn{7}{@{}l}{
            \textit{Qwen3-235B $\rightarrow$ Qwen3-1.7B}
        } \\
        \multirow{2}{*}{PLatour-SFT}
        & Equal-width
        & 89.25 & 77.22 & 79.05 & 50.18 & 82.52 \\
        & Last-$n^{\dagger}$
        & 95.56 & 82.90 & 78.57 & 57.80 & 85.63 \\
        \addlinespace
        \multirow{2}{*}{PLatour-GRPO}
        & Equal-width$^{\dagger}$
        & -- & -- & -- & -- & 89.35 \\
        & Last-$n$
        & -- & -- & -- & -- & 86.50 \\
        \midrule
        \multicolumn{7}{@{}l}{
            \textit{Qwen3-8B $\rightarrow$ Qwen3-1.7B}
        } \\
        \multirow{2}{*}{PLatour-SFT}
        & Equal-width
        & 87.54 & 70.31 & 76.67 & 41.61 & 82.98 \\
        & Last-$n^{\dagger}$
        & 89.85 & 71.29 & 52.86 & 38.69 & 83.62 \\
        \bottomrule
    \end{tabular}
    \caption{Layer-matching ablation for PLatour-SFT and PLatour-GRPO with a Qwen3-1.7B receiver. All scores are percentages, with evaluation settings specified in Section~\ref{app:evaluation}. Unreported results are denoted by --. $^{\dagger}$ marks the configurations used in Table~\ref{tab:qwen3_cross_model_transfer}.}
    \label{tab:layer_matching_ablation}
\end{table}

\begin{table}[H]
    \centering
    \footnotesize
    \setlength{\tabcolsep}{6pt}
    \begin{tabular}{@{}rrr@{}}
        \toprule
        & \multicolumn{2}{c}{avg@4 (\%)} \\
        \cmidrule(l){2-3}
        Step & Last-$n$ & Relative-depth (full-to-full) \\
        \midrule
        0   & 83.65 & 83.25 \\
        30  & 88.60 & 91.25 \\
        59  & 82.95 & 89.15 \\
        89  & 89.20 & 90.00 \\
        118 & 90.75 & 91.10 \\
        \bottomrule
    \end{tabular}
    \caption{Comparison of last-$n$ and relative-depth (full-to-full) layer matching across training steps. All scores are avg@4, reported as percentages.}
    \label{tab:grpo_layer_match}
\end{table}

\subsection{Hidden-State versus KV Transfer}
\label{app:hidden_vs_kv}

Table~\ref{tab:hidden_vs_kv} compares hidden-state and KV transfer using a Qwen3-8B sender and a Qwen3-1.7B receiver. We replace C2C's original projector with linear maps, using either separate or shared projections for keys and values. This change alone does not improve accuracy. Zero initialization substantially improves both variants, bringing them close to, but still below, the standalone receiver. Enabling propagation further improves accuracy by 2.35--2.76 percentage points, allowing both KV-transfer variants to outperform the standalone receiver. Nevertheless, PLatour-SFT remains ahead by 1.25--1.40 points.

One possible advantage of hidden-state transfer is that it learns a single mapping before the receiver constructs its keys and values through its pretrained projections. This couples the transferred information through a common hidden representation, whereas separate KV mappings learn the two transformations independently. To investigate whether sharing projection parameters alone explains the gap, we also tie the key and value projection weights. This produces only small accuracy changes and does not close the gap with PLatour-SFT, suggesting that parameter sharing alone is insufficient to explain the advantage of hidden-state transfer.

\begin{table}[htbp]
    \centering
    \footnotesize
    \setlength{\tabcolsep}{5pt}
    \begin{tabular}{@{}llllrr@{}}
        \toprule
        Method & Projection & Init. & Propagation & Accuracy (\%) & Bridge Weight Norm \\
        \midrule
        Receiver alone & --           & --    & -- & 80.86 & -- \\
        C2C            & Original     & Random  & No & 75.85 & 16.61\\
        \midrule
        Linear C2C     & Separate K/V & Random & No & 75.63 & 18.48\\
        Linear C2C     & Separate K/V & Zero   & No & 79.87 & 1.17\\
        Linear C2C     & Separate K/V & Zero   & Yes & 82.22 & 1.10\\
        \midrule
        Linear C2C     & Shared K/V   & Random & No & 74.68 & 18.48\\
        Linear C2C     & Shared K/V   & Zero   & No & 79.61 & 1.15\\
        Linear C2C     & Shared K/V   & Zero   & Yes & 82.37 & 1.10\\

        \midrule
        PLatour-SFT    & Hidden-state & Zero  & Yes & \textbf{83.62} & 3.60\\
        \bottomrule
    \end{tabular}
    \caption{GSM8K avg@2 accuracy with a Qwen3-8B sender and a Qwen3-1.7B receiver, using a 2,048-token response limit. Shared K/V uses the same linear map for keys and values; separate K/V uses independently parameterized maps. Propagation indicates whether lower-layer fused KV caches influence higher-layer ones.}
    \label{tab:hidden_vs_kv}
\end{table}

These results suggest that neither zero initialization nor K/V weight sharing alone accounts for the advantage of hidden-state transfer in this comparison. One possible explanation is that PLatour communicates a shared hidden representation from which the receiver derives keys and values through its existing projections, providing a more suitable interface than directly mapping the KV caches.

Bridge Weight Norm denotes the arithmetic mean of the Frobenius norms of all learned two-dimensional bridge weight matrices: $\frac{1}{M}\sum_{j=1}^{M}\|W_j\|_F$, where $M$ is the number of such matrices. Biases, gates, and frozen backbone parameters are excluded. Since matrix dimensions differ across architectures, these values are not normalized for parameter count. It shows that random initialisation has a much larger norm, and it may be the central cause for performance degradation even when C2C is applying residual addition for KV fusion.

\subsection{Effect of Projection Initialization}
\label{app:init}

We compare zero and random initialization of the linear projector under direct GRPO training with a Qwen3-30B sender and a Qwen3-4B receiver. Zero initialization produces no initial residual injection, preserving the receiver's behavior before training. Table~\ref{tab:init_trajectory} shows that zero initialization achieves 84.7\% MATH500 avg@4 accuracy at step 0, compared with 79.3\% for random initialization. At the same training budget of 120 steps, zero initialization reaches 88.1\%, outperforming random initialization by 3.0 percentage points. It also exceeds the randomly initialized variant's final accuracy after only 30 steps. These results support zero initialization as the default for PLatour.

\begin{table}[H]
    \centering
    \footnotesize
    \setlength{\tabcolsep}{6pt}
    \begin{tabular}{@{}lccccc@{}}
        \toprule
        & \multicolumn{5}{c}{GRPO step} \\
        \cmidrule(l){2-6}
        Initialization & 0 & 30 & 60 & 90 & 120 \\
        \midrule
        Zero   & 84.7 & 85.6 & 87.9 & 87.5 & \textbf{88.1} \\
        Random & 79.3 & 82.4 & 83.7 & 84.4 & 85.1 \\
        \bottomrule
    \end{tabular}
    \caption{MATH500 avg@4 accuracy (\%) during direct GRPO training with zero and random projector initialization. Step 0 denotes evaluation before training.}
    \label{tab:init_trajectory}
\end{table}

\subsubsection{KV-Lingo Ablations}
\label{app:kvlingo_ablation}

The original KV-Lingo training recipe differs from that of PLatour-SFT in its learning-rate schedule (cosine rather than linear decay) and warmup ratio (5\% rather than 10\%). We first evaluate the effect of adopting the PLatour-SFT recipe on answer-only tasks. As shown in Table~\ref{tab:kvlingo_recipe_ablation}, this change provides no consistent benefit, reducing accuracy in five of the six sender--benchmark combinations. Nevertheless, we retain this recipe for subsequent KV-Lingo ablations to align the optimization settings with those of C2C and PLatour-SFT.

We next replace KV-Lingo's KL-divergence loss with cross-entropy (CE). The KL objective is intended to preserve the receiver's behavior while adapting to the sender's KV caches, whereas CE directly optimizes the likelihood of target responses. Relative to the recipe-only variant, CE improves accuracy on all three answer-only benchmarks for both sender sizes. In the Qwen3-8B $\rightarrow$ Qwen3-1.7B setting, it outperforms both C2C and PLatour-SFT on all three benchmarks. However, the CE variant remains below the receiver-alone baseline on every generation benchmark for both sender sizes, with particularly severe degradation in the Qwen3-235B $\rightarrow$ Qwen3-1.7B setting. We hypothesize that using receiver-generated rollouts as SFT targets may contribute to this limitation, although out ablations do not isolate this factor.

Finally, we examine KV-Lingo's use of direct KV-cache replacement by replacing residual addition in PLatour-SFT with this mechanism, yielding the \textit{PLatour-SFT Replace} variant. Compared with PLatour-SFT, replacement improves accuracy on all three answer-only benchmarks with the 8B sender, but reduces accuracy on ARC-C and Chemistry with the 235B sender, while improving MMLU-Redux. On generation tasks, replacement consistently underperforms PLatour-SFT for both sender sizes, with especially large drops on GSM8K and MATH500 with the 235B sender. We hypothesize that the larger sender--receiver capacity gap makes it harder for the linear-regression initialization to align their KV representations adequately, leaving direct replacement more vulnerable to representation mismatch.

Taken together, these ablations favor CE over KL for answer-only tasks and residual addition over direct KV-cache replacement for generation tasks, with the latter advantage being particularly pronounced when the sender--receiver capacity gap is large.

\begin{table}[htbp]
    \centering
    \footnotesize
    \setlength{\tabcolsep}{5pt}
    \begin{tabular}{@{}lrrrrrr@{}}
        \toprule
        Method / Recipe & ARC-C & MMLU-Redux & Chemistry & Chemistry (Gen) & GSM8K & MATH500 \\
        \midrule
        Receiver-Alone & 74.23 & 53.96 & 39.05 & 41.43 & 80.86 & 73.40 \\
        \midrule
        \multicolumn{7}{@{}l}{
            \textit{Qwen3-235B $\rightarrow$ Qwen3-1.7B}
        } \\
        \textit{C2C}         & 92.75 & 83.68 & 60.95 & 41.01 & 75.09 & 64.60 \\
        \textit{PLatour-SFT} & 95.56 & 82.90 & 78.57 & 57.80 & 85.63 & 70.65 \\
        \addlinespace
        Original            & 91.21 & 77.88 & 40.95 & 40.95 & 17.06 & 11.95 \\
        PLatour Recipe      & 90.19 & 77.70 & 39.05 & -- & -- & -- \\
        PLatour Recipe + CE & 92.92 & 81.02 & 46.19 & 4.64 & 12.93 & 7.30 \\
        PLatour-SFT Replace & 94.11 & 84.55 & 63.81 & 36.70 & 41.39 & 37.50 \\
        \midrule
        \multicolumn{7}{@{}l}{
            \textit{Qwen3-8B $\rightarrow$ Qwen3-1.7B}
        } \\
        \textit{C2C}         & 84.81 & 72.90 & 55.71 & 34.94 & 75.85 & 67.90 \\
        \textit{PLatour-SFT} & 89.85 & 71.29 & 52.86 & 38.69 & 83.62 & 72.10 \\
        \addlinespace
        Original            & 86.69 & 70.17 & 43.33 & 40.48 & 78.35 & 69.85 \\
        PLatour Recipe      & 87.03 & 69.46 & 39.05 & -- & -- & -- \\
        PLatour Recipe + CE & 89.93 & 73.97 & 61.90 & 33.63 & 79.08 & 68.65 \\
        PLatour-SFT Replace & 90.78 & 74.25 & 74.76 & 37.92 & 83.13 & 70.75 \\
        \bottomrule
    \end{tabular}
    \caption{\textbf{KV-Lingo training-recipe ablation on answer-only and generation tasks (accuracy, \%).} Original denotes KV-Lingo with cosine decay and 5\% warmup; PLatour Recipe uses linear decay with 10\% warmup. All other KV-Lingo settings remain unchanged. Italicized methods are included as reference baselines. Unreported results are denoted by --.}
    \label{tab:kvlingo_recipe_ablation}
\end{table}

\FloatBarrier
\section{Full Results across Supervision Sources}
\label{app:sft_supervision}

For Chemistry (CoT), GSM8K, and MATH, each fine-tuning method has three variants using training responses generated by Qwen3-235B, Qwen3-8B, or Qwen3-1.7B. For brevity, Table~\ref{tab:confound_qwen3} reports only the best-performing variant for each method and benchmark. We provide the results for each supervision source here.

\begin{table}[H]
    \centering
    \footnotesize
    \setlength{\tabcolsep}{8pt}
    \begin{tabular}{@{}llrrr@{}}
        \toprule
        Method & Supervision Source
        & Chemistry (CoT) & GSM8K & MATH500 \\
        \midrule
        \multirow{3}{*}{LoRA-SFT}
        & Qwen3-235B & \textbf{48.10} & 81.16 & 70.00 \\
        & Qwen3-8B   & 38.99 & \textbf{81.69} & \textbf{72.55} \\
        & Qwen3-1.7B & 40.83 & 81.08 & 72.05 \\
        \midrule
        \multirow{3}{*}{Full-SFT}
        & Qwen3-235B & \textbf{51.13} & 83.59 & 70.55 \\
        & Qwen3-8B   & 40.65 & 82.56 & \textbf{73.40} \\
        & Qwen3-1.7B & 42.62 & 81.61 & 71.85 \\
        \bottomrule
    \end{tabular}
    \caption{Fine-tuning results (\%) with training responses generated by different Qwen3 models. All variants adapt the Qwen3-1.7B receiver. Bold values indicate the best supervision source within each method and benchmark, corresponding to the variants reported in Table~\ref{tab:confound_qwen3}.}
    \label{tab:lora_sft_appendix}
\end{table}

\section{Projection Norms under SFT and GRPO}
\label{app:norm}

We compare the aggregate Frobenius norms of the learned projection matrices under PLatour-SFT and PLatour-GRPO. The projection norm reveals the scale of the learned bridge, while the actual injection magnitude also depends on the sender activations and their alignment with the projection matrices.

As shown in Table~\ref{tab:frobenius_norm}, SFT produces projection norms that are $4.32$--$8.32$ times those learned by GRPO under our training settings. Larger projection norms may increase sensitivity to replacing question-dependent sender features with features from an unrelated problem, as shown in Table~\ref{tab:confound_qwen3}.

\begin{table}[H]
    \centering
    \footnotesize
    \setlength{\tabcolsep}{6pt}
    \begin{tabular}{@{}lrrr@{}}
        \toprule
        Sender $\to$ Receiver & SFT & GRPO & SFT / GRPO \\
        \midrule
        \multicolumn{4}{@{}l}{\textit{GSM8K}} \\
        235B $\to$ 1.7B & 21.128 & 2.634 & $8.02\times$ \\
        8B $\to$ 1.7B   & 19.132 & 2.299 & $8.32\times$ \\
        \midrule
        \multicolumn{4}{@{}l}{\textit{Chemistry (CoT)}} \\
        235B $\to$ 1.7B & 11.568 & 1.855 & $6.24\times$ \\
        8B $\to$ 1.7B   &  9.400 & 2.178 & $4.32\times$ \\
        \bottomrule
    \end{tabular}
    \caption{Aggregate Frobenius norms of the projection matrices across 28 receiver layers at the final training checkpoints. Each configuration contains 234,881,024 projection parameters.}
    \label{tab:frobenius_norm}
\end{table}

\section{Training Instability and Clipping on MATH and DAPO}
\label{app:clip}

\paragraph{MATH.} We observe training instability in PLatour-GRPO with a Qwen3-235B sender and a Qwen3-1.7B receiver on MATH. As shown in Table~\ref{tab:platour_lora_training}, accuracy initially improves from 73.10\% to 78.20\%, but collapses to 0.00\% at step 177 and remains there at subsequent checkpoints. This collapse is accompanied by elevated clipping rates of 54.75--59.40\%. In comparison, LoRA-GRPO continues to improve, reaching 82.25\% at step 295. These results identify a stability limitation of PLatour in this configuration. Although increased clipping accompanies the collapse, the reported statistics do not establish its cause. We leave a systematic investigation of this failure mode to future work.

\begin{table}[H]
    \centering
    \footnotesize
    \setlength{\tabcolsep}{8pt}
    \begin{tabular}{@{}crrrrrr@{}}
        \toprule
        \multirow{2}{*}{Step}
        & \multicolumn{3}{c}{PLatour-GRPO}
        & \multicolumn{3}{c}{LoRA-GRPO} \\
        \cmidrule(lr){2-4}
        \cmidrule(l){5-7}
        & Accuracy & Avg.\ Length & Clipping
        & Accuracy & Avg.\ Length & Clipping \\
        \midrule
          0 & 73.10\% &   878 &  2.30\%
            & 73.35\% &   872 &  2.45\% \\
         59 & 78.20\% & 1,275 &  9.65\%
            & 78.20\% & 1,269 & 10.50\% \\
        118 & 76.45\% &   992 &  5.55\%
            & 79.60\% & 1,433 & 13.20\% \\
        177 &  0.00\% & 2,438 & 59.40\%
            & 80.10\% & 1,486 & 14.45\% \\
        236 &  0.00\% & 2,246 & 54.75\%
            & 80.60\% & 1,505 & 14.40\% \\
        295 &  0.00\% & 2,418 & 58.90\%
            & 82.25\% & 1,455 & 11.50\% \\
        \bottomrule
    \end{tabular}
    \caption{MATH500 avg@4 accuracy, average response length (tokens), and clipping rate during GRPO training. Clipping denotes responses reaching the 4,096-token generation limit. Both methods use a Qwen3-1.7B receiver; PLatour additionally uses a Qwen3-235B sender.}
    \label{tab:platour_lora_training}
\end{table}

\begin{figure}[H]
  \centering
  \includegraphics[width=\textwidth]{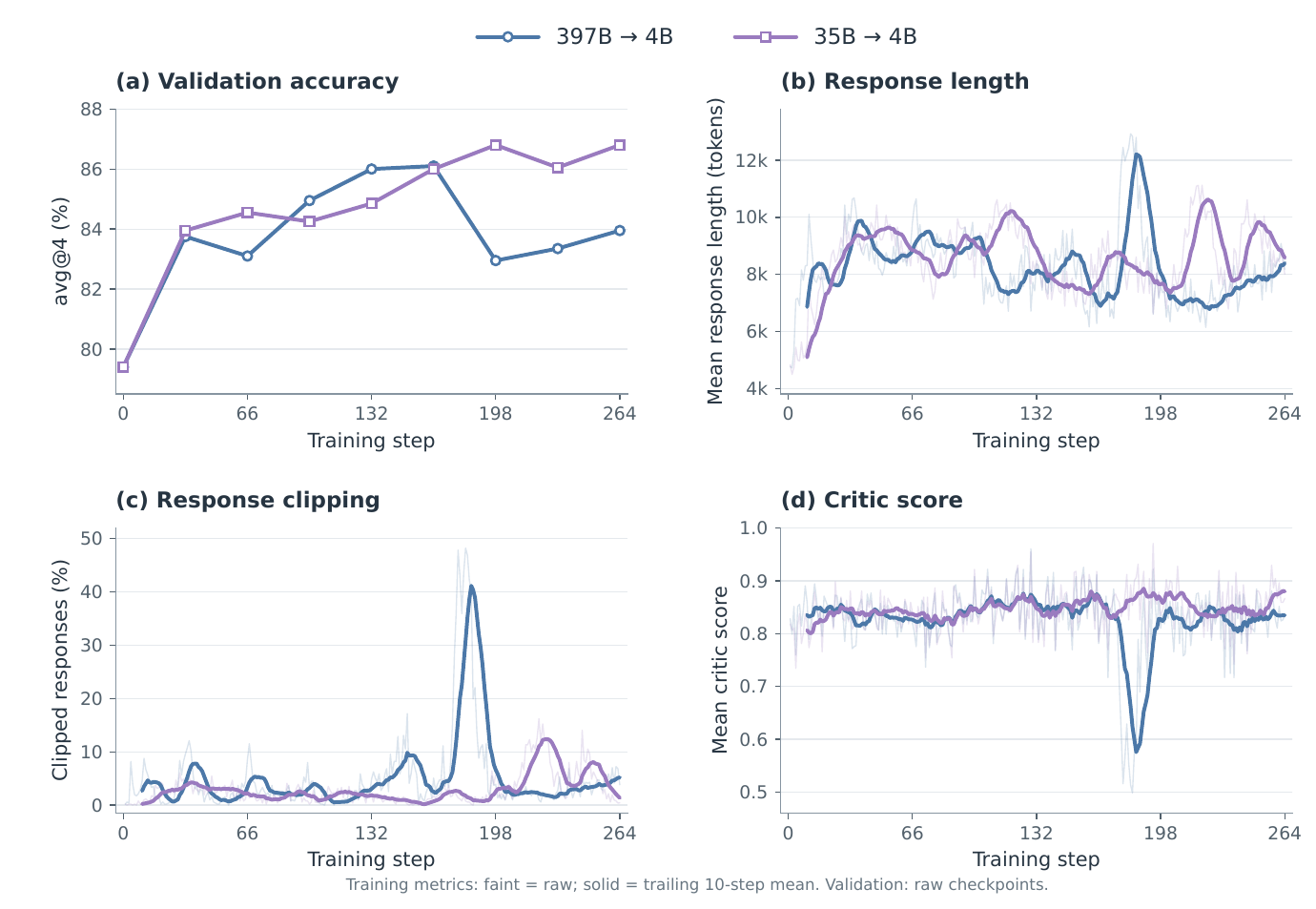}
  \caption{\textbf{Training dynamics of PLatour on DAPO.} We compare Qwen3.5-397B and Qwen3.5-35B senders paired with a Qwen3.5-4B receiver. Panels show (a) validation accuracy (avg@4), (b) mean response length, (c) the percentage of clipped responses, and (d) mean critic score. For training metrics, faint curves show raw values and solid curves show trailing 10-step means; validation accuracy is shown at evaluated checkpoints. The 397B-sender run exhibits a transient clipping spike around step 180, together with increased response length and a sharp drop in critic score. These training metrics subsequently recover, but validation accuracy remains below its earlier peak.}
  \label{fig:dapo_clipping}
\end{figure}

\paragraph{DAPO.}
We also observe training instability when training PLatour on DAPO with a Qwen3.5-397B sender and a Qwen3.5-4B receiver. As shown in Figure~\ref{fig:dapo_clipping}, validation accuracy initially improves to approximately 86\% avg@4 by step 165. Around step 180, the run exhibits a pronounced but transient spike in response clipping, accompanied by a sharp increase in response length and a drop in critic score. The smoothed clipping rate reaches approximately 40\%, while the mean critic score falls from around 0.85 to below 0.60. Although these training metrics subsequently return to their earlier ranges, validation accuracy drops to approximately 83\% at step 198 and recovers only partially, reaching approximately 84\% at step 264. In comparison, the 35B-sender run exhibits milder clipping fluctuations and finishes at approximately 87\% validation accuracy. This episode illustrates a transient training disruption followed by a lasting loss in validation performance, although the plotted metrics alone do not establish response clipping as its cause.

\clearpage
\section{Full Results for PACT-Check}
Tables~\ref{tab:confound_qwen3} and~\ref{tab:confound_qwen3.5} report the full PACT-Check results.

\begin{table}[H]
    \centering
    \footnotesize
    \setlength{\tabcolsep}{2pt}
    \begin{tabular}{@{}ll cccc cccc@{}}
        \toprule
        &
        & \multicolumn{4}{c}{Answer-Only / Without CoT}
        & \multicolumn{4}{c}{Generation / With CoT} \\
        \cmidrule(lr){3-6}
        \cmidrule(l){7-10}
        Setting & Method & ARC-C & MMLU-Redux & Chemistry & Avg.
        & Chemistry & GSM8K & MATH500 & Avg. \\
        \midrule
        \multicolumn{2}{@{}l}{Receiver-Alone}
        & 74.23 & 53.96 & 39.05 & 55.75
        & 41.43 & 80.86 & 73.40 & 65.23 \\
        \midrule
        Cross-Model Transfer & PLatour-SFT
        & 95.56 & 82.90 & 78.57 & 85.68
        & 57.80 & 85.63 & 70.65 & 71.36 \\
        & PLatour
        & -- & -- & -- & --
        & 69.29 & 89.35 & 78.20 & 78.95 \\
        \midrule
        Dummy Transfer (DT) & PLatour-SFT
        & 75.68 & 57.90 & 39.05 & 57.54
        & 38.04 & 80.71 & 70.90 & 63.22 \\
        & PLatour
        & -- & -- & -- & --
        & 52.44 & 87.76 & 78.20 & 72.80 \\
        \addlinespace
        Irrelevant Transfer (IT) & PLatour-SFT
        & 24.57 & 25.09 & 25.24 & 24.97
        & 22.68 & 0.80 & 0.60 & 8.03 \\
        & PLatour
        & -- & -- & -- & --
        & 62.68 & \underline{89.12} & 78.55 & 76.78 \\
        \addlinespace
        Trained Receiver (TR) & Full-SFT 
        & \underline{81.06} & 60.81 &  77.62 & 73.16 & 51.13 & 83.59 & 73.40 & 69.37\\
        & LoRA-SFT
        & 80.72 & \underline{61.54} & 77.62 & 73.29
        & 48.10 & 81.69 & 72.55 & 67.45 \\
        & Full-GRPO
        & -- & -- & -- & -- & \underline{65.95} & 88.02 & 81.50 & 78.49 \\
        & LoRA-GRPO
        & -- & -- & -- & --
        & 62.92 & \underline{88.53} & \underline{82.25} & 77.90 \\

        \addlinespace
        Self-Communication (SC) & PLatour-SFT
        & 76.62 & 56.20 & \underline{79.52} & 70.78
        & 40.95 & 81.16 & 71.90 & 64.67 \\
        & PLatour
        & -- & -- & -- & --
        & 60.95 & 88.06 & 80.95 & 76.65 \\
        \addlinespace
        Virtual Sender & PLatour-SFT
        & 50.09 & 51.67 & 73.81 & 58.52
        & 42.92 & 82.03 & 72.65 & 65.87 \\
        \bottomrule
    \end{tabular}
    \caption{Comparison of cross-model communication with control settings (accuracy, \%). All settings use a Qwen3-1.7B receiver. Cross-model rows use a Qwen3-235B sender; Self-Communication uses Qwen3-1.7B as both sender and receiver; Virtual Sender uses trainable virtual sender representations. LoRA-SFT reports the best result for each benchmark across the configurations in Table~\ref{tab:lora_sft_appendix}. Avg.\ denotes the arithmetic mean of the three scores in each group and is reported only when all three scores are available. Underlined scores mark the strongest of Dummy Transfer, Irrelevant Transfer, Trained Receiver, and Self-Communication for PLatour-SFT or PLatour in each column, from which PACT-Score is computed in Table~\ref{tab:pact_score}.}
    \label{tab:confound_qwen3}
\end{table}

\begin{table}[H]
    \centering
    \footnotesize
    \setlength{\tabcolsep}{3.5pt}
    \begin{tabular}{@{}l ccccc cccc@{}}
        \toprule
        & \multicolumn{5}{c}{Mathematics (DAPO-17k)}
        & \multicolumn{4}{c}{Coding (DeepCoder-24k Clean)} \\
        \cmidrule(lr){2-6}
        \cmidrule(l){7-10}
        Method & AIME-80 & AIME-135 & HMMT & IMO-Ans. & Avg.
        & LCB v6 & HumanEval & MBPP & Avg. \\
        \midrule
        Receiver-Alone
        & 59.22 & 59.12 & 46.35 & 41.13 & 48.90
        & 40.36 & 83.16 & 65.08 & 62.87 \\
        \midrule
        \multicolumn{10}{@{}l}{\textit{Cross-model transfer}} \\
        PLatour (35B)
        & 75.63 & 74.63 & 64.58 & 49.75 & 63.32
        & 54.29 & 94.97 & 89.93 & 79.73 \\
        Mostik (753B)
        & -- & 70.74\rlap{$^{*}$} & -- & -- & --
        & 54.14\rlap{$^{*}$} & 86.13\rlap{$^{*}$} & 82.00\rlap{$^{*}$} & 74.09 \\
        \midrule
        \multicolumn{10}{@{}l}{\textit{Controls for PLatour}} \\
        Dummy Transfer
        & 74.84 & 74.91 & 65.10 & 49.50 & 63.15
        & 55.36 & 95.20 & 88.88 & 79.81 \\
        Irrelevant Transfer
        & 74.53 & 74.26 & \underline{65.94} & 49.63 & 63.36
        & 54.05 & 94.33 & 89.50 & 79.29 \\
        Trained Receiver (Full-Param)
        & \underline{76.95} & \underline{77.04} & 65.21 & \underline{51.38} & \underline{64.51}
        & 54.29 & \underline{96.49} & \underline{91.38} & \underline{80.72} \\
        Trained Receiver (LoRA)
        & 72.73 & 73.56 & 59.48 & 47.44 & 59.88
        & \underline{56.29} & 91.46 & 82.83 & 76.86 \\
        Self-Communication
        & 72.42 & 72.41 & 56.98 & 43.69 & 57.70
        & 53.14 & 95.43 & 90.85 & 79.81 \\
        Virtual Sender
        & 72.66 & 72.73 & 59.79 & 44.56 & 59.00
        & 56.00 & 91.77 & 85.95 & 77.91 \\
        \bottomrule
    \end{tabular}
    \caption{Comparison of cross-model communication with control settings (accuracy, \%). All settings use a Qwen3.5-4B receiver. Cross-model PLatour uses a Qwen3.5-35B sender; Mostik uses GLM-5.2 (753B) and is shown for comparison. Trained Receiver fine-tunes the receiver with GRPO, either with full parameters or with LoRA. AIME, IMO-Ans., and Avg.\ are as in Table~\ref{tab:qwen35_res}; the controls and underlines are as in Table~\ref{tab:confound_qwen3}. Our coding experiments use DeepCoder-24k Clean, decontaminated against all three coding test sets. $^{*}$Mostik scores are reported results.}
    \label{tab:confound_qwen3.5}
\end{table}

\section{Full Results for Dummy Transfer and Irrelevant Transfer}
\label{app:sender_control_results}

Table~\ref{tab:confound_qwen3} summarizes the main control comparison via Qwen3-235B sender and Qwen3-1.7B receiver. Table~\ref{tab:sender-to-1.7b} provides expanded per-benchmark results for both Qwen3-235B and Qwen3-8B senders. We report original, dummy-transfer, and irrelevant-transfer results for PLatour-SFT, PLatour-GRPO, C2C, and KV-Lingo, alongside receiver-only baselines and the available self-communication result. All controls follow Section~\ref{app:sender_controls} and use the corresponding original checkpoint without additional training.

\begin{table}[H]
    \centering
    \footnotesize
    \setlength{\tabcolsep}{5pt}
    \begin{tabular}{@{}llrrrrrr@{}}
        \toprule
        & & \multicolumn{3}{c}{Answer-only}
          & \multicolumn{3}{c}{CoT Generation} \\
        \cmidrule(lr){3-5}
        \cmidrule(l){6-8}
        Method & Inference Setting
        & ARC-C & MMLU-Redux & Chemistry
        & Chemistry & GSM8K & MATH500 \\
        \midrule
        Receiver alone (1.7B) & Receiver-only
        & 74.23 & 53.96 & 39.05 & 41.43 & 80.86 & 73.40 \\
        Best trained receiver & Per-task best
        & 81.06 & 61.54 & 77.62 & 65.95 & 88.53 & 82.25 \\
        \midrule
        \multicolumn{8}{@{}l}{
            \textit{Qwen3-235B $\rightarrow$ Qwen3-1.7B}
        } \\
        \multirow{3}{*}{PLatour-SFT}
        & Original
        & 95.56 & 82.90 & 78.57 & 57.80 & 85.63 & 70.65 \\
        & Dummy Transfer
        & 75.68 & 57.90 & 39.05 & 38.04 & 80.71 & 70.90 \\
        & Irrelevant Transfer
        & 24.57 & 25.09 & 25.24 & 22.68 & 0.80 & 0.60 \\
        \addlinespace
        \multirow{3}{*}{PLatour}
        & Original
        & -- & -- & -- & 69.29 & 89.35 & 78.20 \\
        & Dummy Transfer
        & -- & -- & -- & 52.44 & 87.76 & 78.20 \\
        & Irrelevant Transfer
        & -- & -- & -- & 62.68 & 89.12 & 78.55 \\
        \addlinespace
        \multirow{3}{*}{C2C}
        & Original
        & 92.75 & 83.68 & 60.95 & 41.01 & 75.09 & 64.60 \\
        & Dummy Transfer
        & 76.02 & 57.94 & 45.71 & 37.74 & 77.10 & 65.05 \\
        & Irrelevant Transfer
        & 46.84 & 26.74 & 46.67 & 39.64 & 75.78 & 65.30 \\
        \addlinespace
        \multirow{3}{*}{KV-Lingo}
        & Original
        & 91.21 & 77.88 & 40.95 & 40.95 & 17.06 & 11.95 \\
        & Dummy Transfer
        & 74.83 & 53.05 & 35.71 & 35.71 & 43.10 & 22.20 \\
        & Irrelevant Transfer
        & 24.66 & 24.64 & 25.24 & 26.67 & 0.76 & 0.75 \\
        Sender-Alone &  Sender-Alone
        & 96.25 & 84.91 & 56.19
        & 58.04 & 94.35 & 89.70 \\
        \midrule
        \multicolumn{8}{@{}l}{
            \textit{Qwen3-8B $\rightarrow$ Qwen3-1.7B}
        } \\
        \multirow{3}{*}{PLatour-SFT}
        & Original
        & 89.85 & 71.29 & 52.86 & 38.69 & 83.62 & 72.10 \\
        & Dummy Transfer
        & 61.69 & 27.79 & 23.33 & 41.19 & 80.25 & 71.40 \\
        & Irrelevant Transfer
        & 24.66 & 25.04 & 20.48 & 25.24 & 0.87 & 0.70 \\
        \addlinespace
        \multirow{3}{*}{PLatour}
        & Original
        & -- & -- & -- & 68.39 & 87.91 & -- \\
        & Dummy Transfer
        & -- & -- & -- & 49.70 & 87.57 & -- \\
        & Irrelevant Transfer
        & -- & -- & -- & 57.32 & 87.98 & -- \\
        \addlinespace
        \multirow{3}{*}{C2C}
        & Original
        & 84.81 & 72.90 & 55.71 & 34.94 & 75.85 & 67.90 \\
        & Dummy Transfer
        & 75.85 & 58.49 & 44.29 & 35.60 & 79.87 & 69.65 \\
        & Irrelevant Transfer
        & 64.08 & 30.11 & 53.33 & 32.74 & 76.00 & 67.70 \\
        \addlinespace
        \multirow{3}{*}{KV-Lingo}
        & Original
        & 86.69 & 70.17 & 43.33 & 40.48 & 78.35 & 69.85 \\
        & Dummy Transfer
        & 76.19 & 55.95 & 40.48 & 40.95 & 80.74 & 72.95 \\
        & Irrelevant Transfer
        & 25.17 & 24.25 & 26.19 & 26.67 & 0.83 & 0.75 \\
        Sender-Alone & Sender-Alone
        & 91.72 & 74.61 & 40.48 
        & 41.13 & 92.49 & 83.80 \\
        \midrule
        \multicolumn{8}{@{}l}{
            \textit{Qwen3-1.7B $\rightarrow$ Qwen3-1.7B
            (Self-Communication)}
        } \\
        C2C & Original
        & 74.91 & 59.64 & 59.05 & 43.15 & 81.08 & 71.95 \\
        \bottomrule
    \end{tabular}
    \caption{Original, dummy-transfer, and irrelevant-transfer results (\%) with a Qwen3-1.7B receiver. Original denotes standard inference with the trained communication modules. Controls follow Section~\ref{app:sender_controls}. PLatour-GRPO results use the checkpoints selected by mean validation accuracy. DT and IT retain the corresponding original checkpoint without additional training. Unreported values are denoted by --.}
    \label{tab:sender-to-1.7b}
    \label{tab:independent}
\end{table}

\end{document}

%% file: method_v2.tex
\section{Background}

\subsection{Problem Definition}

PACT aims to improve a receiver model's generation performance by incorporating prompt representations from a sender model into the receiver's initial inference state. Given an input prompt $x$, let $H_S(x)$ denote the sender's hidden states obtained during prefilling, and let $\mathcal{B}_{\phi}$ denote a trainable communication bridge parameterized by $\phi$. The receiver's initial state is constructed as
\begin{equation}
    z_R^{\phi}(x)
    =
    \mathrm{Prefill}_R
    \bigl(x;\mathcal{B}_{\phi}(H_S(x))\bigr),
\end{equation}
where the second argument represents latent information injected during sender prefilling. The resulting state $z_R^{\phi}(x)$ includes the receiver's cached attention or recurrent states.

The receiver then generates a response $y$ without further sender involvement:
\begin{equation}
    p_{\mathrm{PACT}}(y \mid x; \phi)
    :=
    \prod_{t=1}^{|y|}
    p_R\bigl(y_t \mid y_{<t}, z_R^\phi(x)\bigr).
    \label{eq:pre-fill-define}
\end{equation}

For a given prompt $x$, our goal is to learn bridge parameters $\phi$ that transfer useful information from the sender into the receiver's state, yielding a higher expected task reward than the standalone receiver:
\begin{equation}
\label{eq:objective}
    \mathbb{E}_{y \sim p_{\mathrm{PACT}}(\cdot \mid x; \phi)}
    [r(x,y)]
    >
    \mathbb{E}_{y \sim p_R(\cdot \mid x)}
    [r(x,y)],
\end{equation}
where $r(x,y)$ is the task reward and $p_R(\cdot \mid x)$ denotes the standalone receiver's generation distribution without sender communication.

\subsection{Shared Representation Hypothesis}

The Platonic Representation Hypothesis \citep{plato} suggests that independently trained models may develop similar relational structures in their representations, despite differences in architecture, training data, and capacity. OPRD \citep{oprd} further reports high cosine similarity between models' hidden-state representations after projection into a shared low-rank space. Together, these findings motivate our hypothesis that representations at corresponding sender and receiver layers share a common structure that can facilitate cross-model latent communication.

Such structural similarity, however, does not guarantee effective information transfer. The sender must provide task-relevant information in a form that the receiver can exploit during generation. We therefore view shared representational structure as a starting point for designing communication bridges, while treating their ability to transfer useful information as an empirical question.

\section{Prefill-Only Latent Communication for Reasoning}

\begin{figure*}[t]
    \centering
    \includegraphics[width=\textwidth]{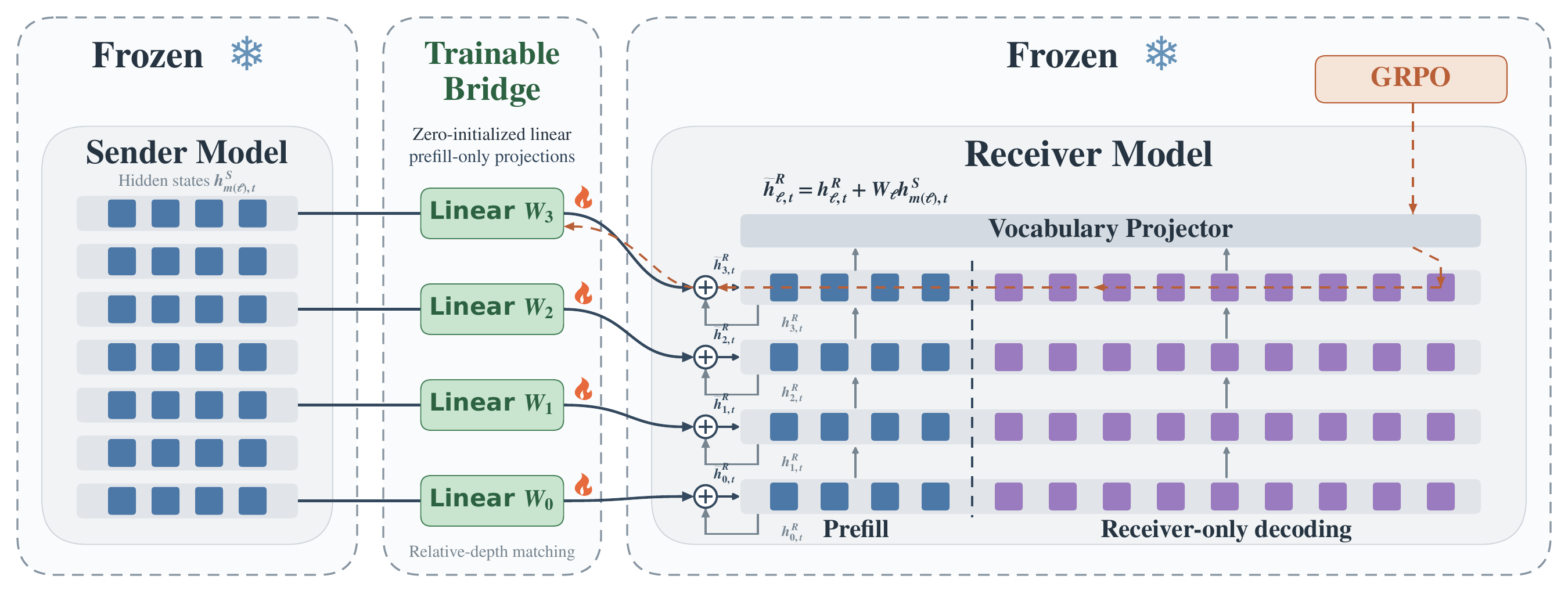}
    \caption{
       \textbf{Overview of PLatour}. During prefilling, independently parameterized, zero-initialized linear projectors $W_i$ map hidden states from matched sender layers into the receiver's residual stream via additive fusion. The receiver subsequently generates the response without further sender involvement. Both model backbones remain frozen; only the projectors are optimized using GRPO. Orange dashed arrows schematically illustrate backpropagation through the frozen receiver to the trainable projectors. }
    \label{fig:PLatour}
\end{figure*}

To study PACT systems on a larger scale and across a broader range of tasks, we propose a strong PACT baseline, PLatour (\textbf{P}refill-only \textbf{LAT}ent c\textbf{O}mm\textbf{U}nication for \textbf{R}easoning). PLatour connects a larger sender model to a smaller receiver model through trainable layer-wise bridges that transfer prompt representations during prefilling (Figure \ref{fig:PLatour}). We view the sender's hidden states as latent tools that can help the receiver, and the bridge learns how to use them effectively. We train the bridges directly with reinforcement learning (RL) to optimize the receiver's use of these representations for downstream generation. Both sender and receiver remain frozen, with only the bridge being updated. We empirically justify our design in Appendix \ref{sec:appendix}.

\subsection{Bridge Architecture}

\paragraph{Linear Projection.}

For each receiver layer $\ell$, we introduce an independent bias-free linear projector $W_\ell \in \mathbb{R}^{d_R \times d_S}$, where $d_S$ and $d_R$ denote the sender and receiver hidden dimensions. Let $h^S_{m(\ell),t}$ and $h^R_{\ell,t}$ denote the hidden states entering the corresponding sender and receiver blocks at token position $t$. We inject the projected sender representation into the receiver's residual stream through additive fusion:
\begin{equation}
\label{eq:hidden}
    \widetilde{h}^R_{\ell,t}
    =
    h^R_{\ell,t}
    +
    a_t W_\ell h^S_{m(\ell),t},
\end{equation}
where $a_t=1$ for all prompt tokens except the last one, and $a_t=0$ otherwise. Communication is therefore restricted to the prefill stage.

We adopt residual addition \citep{c2c} rather than replacing the receiver's hidden states with projected sender representations. Zero initialization preserves the standalone receiver's computation before training. Its benefits are supported by the direct-GRPO initialization trajectories in Table~\ref{tab:init_trajectory} and the SFT comparisons in Table~\ref{tab:hidden_vs_kv}. LoRA \citep{lora} adopts the same initialization principle. Direct replacement does not provide this initialization property and may require an additional alignment stage.

We use linear projectors instead of nonlinear MLPs to simplify the communication mechanism and facilitate its analysis. Linear projectors also achieve better performance in our ablation study (Table \ref{tab:ablation}), potentially because they are easier to optimize. By transferring hidden states rather than directly mapping KV caches, this design allows PLatour to accommodate both standard attention and hybrid architectures with recurrent states.

\paragraph{Propagation.} 

Bridge-modified hidden states are processed by the receiver and propagated to subsequent layers. This preserves the receiver's sequential information flow, allowing higher layers to process representations that already reflect lower-layer injections. Rather than modifying independently computed states at each layer, this design incorporates communication into the receiver's ongoing computation, keeping higher-layer states consistent with the interventions applied at lower layers.

\paragraph{Layer Correspondence.}
We establish sender--receiver correspondence at the layer level based on their relative depths. Let $L_S$ and $L_R$ denote the numbers of sender and receiver layers, respectively. Using zero-based indexing, receiver layer $\ell$ is initially paired with sender layer
\begin{equation}
    \begin{aligned}
        m_0(\ell) &= \operatorname{round}\left( \frac{\ell(L_S - 1)}{L_R - 1} \right), \\
        \ell &= 0, \ldots, L_R - 1.
    \end{aligned}
\end{equation}
This mapping approximately preserves normalized layer depth across models of different sizes.

For hybrid-attention architectures, we additionally impose an asymmetric attention-type constraint. A linear-attention receiver layer can be paired with either a linear- or full-attention sender layer, whereas a full-attention receiver layer must be paired with a full-attention sender layer. If the initial mapping violates this constraint, we select the nearest full-attention sender layer. Motivated by the intuition that linear attention has a stronger locality bias, we pair full-attention receiver layers only with full-attention sender layers to favor the transfer of global contextual information.

\subsection{Training Objective}

We freeze sender and receiver and optimize only the communication bridge using GRPO \citep{grpo}, without preceding SFT alignment or distillation warm-up. PLatour directly maximizes the receiver's expected task reward, $\mathbb{E}_{y \sim p_{\mathrm{PACT}}(\cdot \mid x; \phi)}[r(x,y)]$. We do not explicitly align sender and receiver hidden representations, since representational agreement alone does not ensure the preservation of task-relevant information. As an extreme example, an unconstrained two-sided alignment objective admits a trivial solution in which both projected representations collapse to zero, achieving perfect agreement while conveying no extra information from the sender. We also avoid training the bridged receiver to imitate the sender's output distribution. A large capacity gap may make such imitation difficult or may even degrade the receiver's performance, consistent with reported challenges in large-to-small on-policy distillation \citep{rethink_opd}. Instead, we view the sender's hidden states as latent tools and train the bridge to make these representations useful for maximizing the receiver's generation reward.

\section{PACT-Check}
\label{sec:pact_check}

PLatour optimizes its bridge using task rewards without explicitly requiring the receiver to use sender information. Although both model backbones remain frozen, the trainable bridge can modify the receiver's computation. Gains over the standalone receiver may therefore arise from useful information transfer, receiver-side adaptation, or both. To investigate these contributions, we introduce PACT-Check, an evaluation protocol comprising Dummy Transfer (DT), Irrelevant Transfer (IT), Self-Communication (SC), and Trained Receiver (TR). DT and IT are inference-only interventions that test a trained system's dependence on input-specific sender information. SC and TR require separate training runs and assess whether gains can be achieved through adaptation without an external sender. More details are in Appendix \ref{app:sender_controls}.

\paragraph{Dummy Transfer (DT).}
DT retains only fixed template-level sender features, removing question-dependent communication. It tests how much of the original gain persists without input-specific sender information. Retaining comparable gains suggests that question-specific sender representations may not be necessary for the observed improvement. Instead, the bridge may act as a learned adaptation mechanism, potentially conditioning its effects on the receiver's own representations.

\paragraph{Irrelevant Transfer (IT).} 
IT extends DT by injecting sender representations of tokens from an unrelated question, while the receiver continues to process the original question. It tests whether performance depends on the relevance of the transferred content. A substantial drop indicates sensitivity to mismatched sender information, although such sensitivity alone does not establish semantic understanding or useful information transfer.

\paragraph{Self-Communication (SC).}
SC uses the receiver model as both sender and receiver and trains the communication modules. When the transferred signal is the receiver's current hidden state, with identity layer matching, Eq.~\ref{eq:hidden} reduces to
\begin{equation}
\label{eq:self_comm}
    \widetilde{h}^R_{\ell,t}
    =
    \bigl(I + a_t W_\ell\bigr)h^R_{\ell,t}.
\end{equation}
In this case, the bridge acts as a linear residual adapter, connecting SC to parameter-efficient fine-tuning \citep{peft}. SC therefore tests whether comparable gains can be obtained through the communication modules' adaptation capacity without a larger sender.

\paragraph{Trained Receiver (TR).}
TR directly adapts the receiver using LoRA or full-parameter fine-tuning with SFT or GRPO. It provides a reference for the gains achievable through receiver-only adaptation under these standard training protocols.

\paragraph{PACT-Score.}
We define PACT-Score as the accuracy margin of a PACT system over its strongest applicable PACT-Check control:
\begin{equation}
\mathrm{PACT\text{-}Score}
=
A_{\mathrm{PACT}}
-
\max_{c \in \mathcal{C}} A_c,
\label{eq:pact_score}
\end{equation}
where $A_{\mathrm{PACT}}$ is the PACT system's accuracy, $A_c$ is the accuracy of control $c$, and $\mathcal{C}$ contains the applicable controls among DT, IT, SC, and TR. A positive score provides evidence for a benefit from sender information. A non-positive score indicates that at least one control matches or exceeds the PACT system's performance, so its gains cannot be uniquely attributed to useful sender information.